\documentclass[onecolumn]{article}
\usepackage[latin1]{inputenc}
\usepackage{amsmath}
\usepackage{amsfonts}
\usepackage{amssymb}
\usepackage{color}
\usepackage{graphicx}
\usepackage{bm}
\usepackage{algorithm,algorithmic}
\usepackage{multicol,multirow}
\usepackage{subfigure}
\usepackage{enumerate}
\usepackage{appendix}

\DeclareMathOperator{\EV}{\mathrm{E}}  % expectation operator
\newtheorem{thm}{Theorem}[section]
\newtheorem{prop}{Proposition}[section]
\newtheorem{lmm}{Lemma}[section]
\newtheorem{prf}{Proof}[section]

\newtheorem{defi}{Definition}[section]

\title{Measures of Entropy from Data Using Infinitely Divisible Kernels}
\author{Luis G. {Sanchez Giraldo}\dag \and Murali Rao\ddag~
        \and Jose C. Principe\dag\\ 
\dag Department of Electrical and Computer Engineering,\\
\ddag Department of Mathematics\\
University of Florida,, Gainesville, FL, 32611 USA. 
\\
sanchez@cnel.ufl.edu, ,mrao@math.ufl.edu, principe@cnel.ufl.edu.}% <-this % stops a space

\begin{document}

\maketitle

\begin{abstract}
Information theory provides principled ways to analyze different inference and learning problems such as hypothesis testing, clustering, dimensionality reduction, classification, among others. However, the use of information theoretic quantities as test statistics, that is, as quantities obtained from empirical data, poses a challenging estimation problem that often leads to strong simplifications such as Gaussian models, or the use of plug in density estimators that are restricted to certain representation of the data. In this paper, a framework to non-parametrically obtain measures of entropy directly from data using operators in reproducing kernel Hilbert spaces defined by infinitely divisible kernels is presented. The entropy functionals, which bear resemblance with quantum entropies, are defined on positive definite matrices and satisfy similar axioms to those of Renyi's definition of entropy. Convergence of the proposed estimators follows from concentration results on the difference between the ordered spectrum of the Gram matrices and the integral operators associated to the population quantities. In this way, capitalizing on both the axiomatic definition of entropy and on the representation power of positive definite kernels, the proposed measure of entropy avoids the estimation of the probability distribution underlying the data. Moreover, estimators of kernel-based conditional entropy and mutual information are  also defined. Numerical experiments on independence tests compare favourably with state of the art.    
\end{abstract}

%--------------------------------------------------------------------------------------------------------------------------------------
\section{Introduction}
%--------------------------------------------------------------------------------------------------------------------------------------
Operational quantities in information theory such as entropy are defined based on the probability laws underlying the data generation process of the system under analysis. Therefore, it is required that the probabilities of the set of possible events regarding the process of interest are given. When these probabilities are not known in advance and the only information available is given by a finite set of samples $\{z_i\}_{i=1}^N$, the use of information theoretic quantities relies on an appropriate estimation process. The so called \textit{``plug in''} estimation approach breaks this task into two steps. First, the data is employed to fit a model of its distribution. Then, this model is plugged into the definition of the information theoretic quantity to convey an estimator. Plug in estimation is intuitive and in some cases straightforward in terms of its computation. Nevertheless, in the case of entropy it relies on the quality of the estimation of the underlying distribution. Estimating the distribution is by itself a difficult problem. For example, for a continuous random variable $X$, using the empirical distribution $\frac{1}{N}\sum_{i=1}^N \delta_{z_i}(x)$ for a plug in estimator is rather meaningless. Parametric models can be employed, but then one is faced with the problem of choosing the appropriate model. Tractability versus making oversimplifying assumptions on the model is thus an important issue. On the other hand, using non-parametric estimators of the data distribution may require tuning of additional hyper-parameters, which can lead to computationally expensive procedures. The possibility of over fitting brings the need to incorporate smoothing and other capacity control mechanisms into the models.\\ 
Despite the above mentioned difficulties, Renyi's definition of entropy along with Parzen density estimation have been successfully applied to learning problems. Suitable versions of information theoretic quantities such as entropy and relative entropy can serve as objective functions \cite{JPrincipe} for a family of unsupervised and supervised learning algorithms. Renyi's entropy of order $\alpha$ of a random  variable $X$  is defined as \cite{ARenyi60},
\begin{equation}\label{eq:Renyi_entropy}
H_{\alpha}(X)  = \frac{1}{1 - \alpha} \log_2{\sum\limits_{x \in \mathcal{X}}p^{\alpha}(x)},
\end{equation} 
where $p$ is the probability mass function, or the probability density function (if $X$ is a continuous random variable the sum becomes an integral when it exists) of the random variable $X$, and $\mathcal{X}$ is the support. From equation \eqref{eq:Renyi_entropy}, we can see that the parameter $\alpha>0$ provides a family of entropy functionals, for which Shannon's entropy corresponds to the case $\alpha \rightarrow 1$. Notice that, for a fixed $\alpha$, if one is interested in comparing distributions by their entropies, the argument of the $\log$ function in  \eqref{eq:Renyi_entropy} would suffice for such comparison. In particular, for the case of $\alpha>1$, the argument of the $\log$ function corresponds to the expected value $\EV[g(p(X))]$, where $g(\cdot)$ is the non-negative monotonically increasing function $g(y) = y^{\alpha-1}$. \\
For $\alpha = 2$, a rather simple yet elegant plug-in estimator of \eqref{eq:Renyi_entropy} can be derived using the Parzen window approximation. Let $\{x_i\}_{i=1}^n \subset \mathbb{R}^d$ be an \emph{i.i.d.} sample of $n$ realizations of the random variable $X$, with density $f(x)$. The Parzen density estimator, $\hat{f}(x) = \frac{1}{n}\sum_{i=1}^{n}\kappa_{\sigma}(x,x_i)$, using a Gaussian kernel $\kappa_{\sigma}(x,y) = C\exp{\left(-\frac{1}{2\sigma^2}\|x-y\|^2\right)}$, where $C$ is a normalization constant and width $\sigma$, which can be plugged into the following integral, yields:
\begin{equation}\label{eq:second_renyi_entropy}
\begin{split}
\hat{H}_2(X) &=  -\log{\int_{\mathcal{X}}\hat{f}^2(x)\mathrm{d}x} \\
	&=-\log{\frac{1}{n^2}\sum\limits_{i,j=1}^n \kappa_{\sqrt{2}\sigma}}(x_i,x_j).
\end{split}
\end{equation}
%A similar expression \eqref{eq:second_renyi_entropy}, but using the kernel size $\sigma$, arises from computing the empirical expectation $\EV_{\textrm{emp}}[\hat{f}(X)]$.
The argument of the $\log$ function in \eqref{eq:second_renyi_entropy} has been called the information potential in analogy to potential fields arising in physics \cite{JPrincipe}. The information potential can be shown to be a special case of a positive definite kernel called the cross information potential that maps probability density functions that are in $L^2$ to a Reproducing kernel Hilbert Space (RKHS) of functions \cite{JWXu08}. This idea has been already exploited to solve optimization problems with information theoretic objective functions that bear close resemblance to kernel methods \cite{LSanchezGiraldo11}. Formulating the problem in terms of the RKHS brings an interesting interpretation of the entropy estimator that goes beyond the obvious relation that is established from employing Parzen windows. The interpretation is that a measure of entropy can be obtained by directly applying a positive definite kernel to the data without having to consider the intermediate step of density estimation. Moreover, it can be shown that the convergence to a population based quantity of the kernel estimators is of order $\mathcal{O}(n^{-1/2})$ and that it is independent of the input dimensionality of $\mathcal{X}$. 

It is important to mention that other approaches based on \textit{entropic graphs} that do not rely upon distribution estimators have been also developed recently. These \textit{``direct''} approaches are based on minimum spanning trees or the travelling salesman problem \cite{AHero98, AHero06} and consistently estimate Renyi's $\alpha$-entropy directly from data samples in $\mathbb{R}^d$. Nevertheless, many of these graph theoretic methods are restricted to entropy estimates for $\alpha \in (0,1)$. Similar work by \cite{NLeonenko08} introduces an asymptotically unbiased estimator of Rényi and Tsallis entropy for all orders $\alpha > 0$ based on the $k$-nearest neighbour distance. A generalization of the $k$-nearest-neighbour method using $k$-nearest-neighbour graphs was introduced by \cite{BPoczos10}. This estimator is consistent for $\alpha \in (0,1)$ with a high probability bound on the estimation error, provided that the entropy estimates correspond to a density that is Lipschitz. These bounds depend on the dimensionality of the domain of the distribution. Despite their nice convergence properties, the above estimators are in general not differentiable and thus their use as objective functions cannot be combined with conventional gradient for optimization. One exception that proposes a differentiable quantity is the  work in \cite{LFaivishevsky08}, which introduces an smooth estimator of Shannon's differential entropy ($\alpha = 1$).
 
In the present work, instead of estimating the probability distribution from the data, we define functionals on normalized positive definite matrices that fulfil similar axiomatic properties of Renyi's $\alpha$ measure of entropy without assuming that probabilities of events are known or have been estimated. The matrices are obtained by evaluating a positive definite kernel on all pairs of data points, which implicitly maps the data to a reproducing kernel Hilbert space of functions. Even though the matrix functionals we present here resemble to well-known definitions in quantum information theory, our approach differs in analysis and scope. In the quantum mechanical setting, the density matrix (operator) describes a mixture of states that the system may assume. In our context, the object of study is the Gram matrix of pairwise evaluations of a positive definite kernel. Following the statistical learning setting where the only available information is contained in a finite \emph{i.i.d.} sample $Z = \{z_i\}_{i=1}^n$, our data-driven entropy acts as a measure of the lack (uncertainty) or presence (structure) of statistical regularities in a given sample represented by a Gram matrix. In the analysis of the information theoretic properties of the proposed functional, we show that the choice of kernel is also key in defining measures of information directly from data. In particular, we show that infinitely divisible kernels (subject to normalization) are well suited for the purpose of obtaining a measure of entropy directly from data.\\
The paper is organized as follows. We start by informally motivating the idea of using a Gram matrix to compute measure of entropy by highlighting the relation between plug in estimation of Renyi's second order entropy based on Parzen windows and the computation of expectation of a an observable in quantum mechanics that employs the concept of density operator. Then, we define the entropy functional on positive definite matrices and show how this functional satisfies a set of axioms that closely follow Renyi's definition of a measure of entropy. Next, a definition of joint entropy using Hadamard products is introduced. 
%This notion brings the representation element into consideration. Namely, w
We show that infinitely divisible matrices are particularly well suited for our definition of entropy and allow a definition of mutual information based on a sum of entropies. To study the asymptotic behaviour of the matrix based measures of entropy, we look at the statistical properties of the Gram matrices based on the relation to the integral operators that arise from positive definite kernels. Furthermore, we provide concentration bounds on the convergence of the eigenvalues of these operators which are independent of the input dimensionality. Finally, we carry out numerical experiments to test independence that compare favourably with the state of the art.
%The functionals From the axiomatic characterization of entropy that leads to Rényi's definition, we develop an analogue version of this function that is applied to positive definite matrices. Then, we look at some basic inequalities of information and how they can be translated to the setting of positive definite matrices. The purpose of this characterization is to establish some desirable properties on the positive definite kernels that can be employed to construct the Gram matrix for which our extension of entropy makes sense in terms of information. 

%--------------------------------------------------------------------------------------------------------------------------------------
\section{Hilbert Space Representation of Data}\label{sec:Hilbert_representation_data}
%--------------------------------------------------------------------------------------------------------------------------------------
The use of reproducing kernel Hilbert spaces to represent data in statistical learning is not a new idea \cite{MAizerman64_perceptron,MAizerman64_klms}. However, it was not until the last two decades that the use of RKHSs became popular in machine learning under the name of kernel methods. 
%Kernel methods provide an appealing framework to deal with data of different nature by embedding abstract set in reproducing kernel Hilbert spaces, where it is possible to carry out manipulations of the representation of data by the operations of addition, scalar multiplication and inner product. 
One of the appeals of this approach is the ability to deal with algorithms in a rather generic way provided the kernel is well-fitted to the particular problem. This property has been recently exploited in many practical applications where data points are not necessarily vectors in $\mathbb{R}^p$, for example in problems involving text, trees, point processes, functional data, among others \cite{JShaweTaylor}.
It has been noticed that kernel induced mappings can be understood as means for computing high order statistics of the data that can  be manipulated in a linear fashion just as first order statistics. Methods such as kernel independent component analysis \cite{FBach02}, the work on measures of dependence and independence using Hilbert-Schmidt norms \cite{AGretton05}, Hilbert space embeddings of probability measures \cite{BSriperumbudur08}, and recent work on quadratic measures of independence \cite{SSeth11} are just among the examples of this emerging line of work. 
%\subsection{Hilbert Space Representation of Data}
To motivate the use of positive definite matrices as suitable descriptors of data, we need to understand the role of the Hilbert space representation and how it naturally arises from basic concepts in pattern recognition. Let $(\mathcal{X},\mathcal{B}_{\mathcal{X}})$ be the object space with a countably generated $\sigma$-algebra $\mathcal{B}_{\mathcal{X}}$ and a probability measure $P_{\mathcal{X}}$ defined on it. A measurable function $\phi:\:\mathcal{X} \mapsto \mathbb{R}$ is called a feature. A representation is a family of features $\{\phi_t \}_{t \in \mathcal{T}}$, where $(\mathcal{T},\mathcal{B}_{\mathcal{T}},\mu_{\mathcal{T}})$ is a measure space and $\mu_{\mathcal{T}}$ is $\sigma$-finite. Let $\phi_t$ be also bounded for all $t \in \mathcal{T}$, and let us denote $\phi_{t}(x)$ by $\phi(t,x)$ where $t \in \mathcal{T}$ and $x \in \mathcal{X}$. If we also require that for all fixed $x$ and $y$ in $\mathcal{X}$, 
\begin{equation}\label{eq:generalized_convolution_representation}
G(x,y) = \int\limits_{\mathcal{T}}\phi(t,x)\phi(t,y)\mathrm{d}\mu_{\mathcal{T}}(t) < \infty.
\end{equation}
Then, the space $\mathcal{F}$ defined as the completion of the set of functions $F$ of the form \footnote{Even though it is not explicitly stated, we assume the construction of a linear space of real functions with domain $\mathcal{T}$},
\begin{equation}\label{eq:linear_span}
F(t) = \sum\limits_{i=1}^N \alpha_i\phi(t,x_i),
\end{equation}
where $\alpha_i \in \mathbb{R}$, $x_i \in \mathcal{X}$, and $\forall N \in \mathbb{N}$, is a Hilbert space representation of the set $\mathcal{X}$. In practice however, dealing explicitly with such an $\mathcal{F}$ may be difficult or even impossible. The following result gives an alternative way to dealing with the Hilbert space representation based on the bivariate function $G(x,y)$ defined in \eqref{eq:generalized_convolution_representation}. Consider the set of functions on $\mathcal{X}$ of the form
\begin{equation}\label{eq:kernel_linear_span}
f(x) = \sum\limits_{i=1}^N \alpha_i G(x,x_i),
\end{equation}
where $\alpha_i \in \mathbb{R}$, $x_i \in \mathcal{X}$, and $\forall N \in \mathbb{N}$. Let us define the inner product between elements $f(x) = \sum_{i=1}^N \alpha_i G(x,x_i)$ and $g(x) = \sum_{j=1}^M \beta_i G(x,x_j)$ of the above set as:
\begin{equation}\label{eq:kernel_inner_product}
\langle f,g\rangle = \sum\limits_{i=1}^N\sum\limits_{j=1}^M \alpha_i\beta_j G(x_i,x_j),
\end{equation}
the completion of the above set is a Hilbert space $\mathcal{H}$ of functions on $\mathcal{X}$. Moreover, $\mathcal{H}$ is a reproducing kernel Hilbert space with kernel $G$. Notice that for any finite set 
$\{x_i\}_{i=1}^N$ we have that
\begin{equation}\label{eq:positive_definite_kernel}
\sum\limits_{i=1}^N\sum\limits_{j=1}^N \alpha_i\alpha_j G(x_i,x_j) \geq 0,
\end{equation}
for all $\bm{\alpha} \in \mathbb{R}^N$. Functions satisfying the above condition are called positive definite. 
\begin{thm}\emph{(Basic Congruence Theorem \cite{EParzen59}):} Let $\mathcal{H}_1$ and $\mathcal{H}_2$ be two abstract Hilbert spaces. Let $\mathcal{X}$  be an index set. Let $\{F(x), \:x \in \mathcal{X} \}$, be a family of vectors which span $\mathcal{H}_1$. Similarly, let $\{f(x), \:x \in \mathcal{X} \}$ be a family of vectors which span $\mathcal{H}_2$. Suppose that, for every $x$ and $y$ in $\mathcal{X}$,
\begin{equation}\label{eq:basic_congruence}
\langle F(x),F(y) \rangle_1 = \langle f(x),f(y) \rangle_2
\end{equation}  
Then the spaces $\mathcal{H}_1$ and $\mathcal{H}_2$ are congruent, and one can define a congruence $\Psi$ from $\mathcal{H}_1$ to $\mathcal{H}_2$ which has the property that $\Psi(F(x)) = f(x)$ for $x \in \mathcal{X}$.
\end{thm}

\begin{prop} \label{prop:congruence_rkhs} Let $\mathcal{X}$ be a compact space. The spaces $\mathcal{F}$ and $\mathcal{H}$ are congruent.
\begin{prf} %\emph{(Proof of proposition \ref{prop:congruence_rkhs}).} 
The congruence follows from the definition of $\mathcal{F}$ and $\mathcal{H}$. For $F = \sum_{i=1}^n \alpha_i \phi(t,x_i)$ simply take $\Psi:\mathcal{F} \mapsto \mathcal{H}$
as:
\[
\begin{split}
\Psi(F)(\cdot) & = \sum\limits_{i=1}^n \alpha_i \int_{\mathcal{T}} \phi(t,x_i)\phi(t,\cdot) \mathrm{d}\mu_{\mathcal{T}}(t) \\
& = \sum\limits_{i=1}^n \alpha_i G(\cdot, x_i) = f(\cdot)
\end{split}
\]
\flushright $\square$
\end{prf}
\end{prop}
The above proposition allows us to perform the analysis of the representation of $\mathcal{X}$ on the equivalence classes that can be formed by using the function $G$ to define relations between the elements of $\mathcal{X}$. From the congruence, we can define a distance function between the representations of two elements $x,y \in \mathcal{X}$ using the function $G$ as follows:
\begin{equation}\label{eq:induced_metric_hilbert_representation}
d^2(\phi(t,x),\phi(t,y)) = G(x,x)+G(y,y)-2G(x,y),
\end{equation}
for convenience we write $d^2(\phi(t,x),\phi(t,y))$ as $d^2(x,y)$. That $d(x,y)$ is indeed a semi-metric is verified in the appendix.  
As we shall see in the following sections, the Gram matrix also plays a fundamental role in establishing the connection between information theoretic concepts and kernel methods.\\

\subsection{An Information Theoretic Example: The Cross-Information Potential RKHS}
The argument of the $\log$ function in the estimator of Renyi's second order entropy based on Parzen windows \eqref{eq:second_renyi_entropy} can be interpreted as a data dependent transformation that transfers the statistical properties of the data distribution to a reproducing kernel Hilbert space. For a window function $h(x,y)$, the convolution integral $\kappa(x,y) = \int_\mathcal{X}h(x,z)h(y,z)\mathrm{d}z$, where $\mathcal{X} \subseteq \mathbb{R}^d$, defines a positive definite function $\kappa$ and therefore a reproducing kernel Hilbert space $\mathcal{H}$. The one to one correspondence between positive definite kernels and reproducing kernel Hilbert spaces established in \cite{NAronszajn50} allows us to define an implicit mapping $\phi:\mathcal{X} \mapsto \mathcal{H}$ such that $\kappa(x,y) = \langle \phi(x), \phi(y)\rangle$. The estimator of the quantity inside the $\log$ called information potential corresponds to the squared norm of $\frac{1}{n}\sum_{i=1}^n \phi(x_i)$, which by the law of large numbers converges to $\Vert \EV[\phi(X)] \Vert^{2}$. Notice, the above norm depends on the mapping $\phi$, which is induced by the kernel $\kappa$ between data samples. Now, let us introduce a general form of the information potential that maps square integrable probability density functions to a RKHS. Let $\mathcal{F}$ bet the set of probability density functions on $\mathbb{R}^d$ that are square integrable. We can define the \textit{cross-information potential} $\mathcal{V}$ (CIP) as a bilinear form that maps densities $f_i,f_j \in \mathcal{F}$ to the real numbers via the integral,
\begin{equation}
\mathcal{V}(f_i,f_j) = \int\limits_{\mathbb{R}^n}f_i(x)f_j(x) \mathrm{d}x
\end{equation}
It is easy to show that for a basis of uniformly bounded, square integrable probability density functions, $\mathcal{V}$ defines a RKHS on the $\mathrm{span}\{\mathcal{F}\}$ (up to completion)\cite{JWXu08}. Now consider the set $\mathcal{G} := \{g = \sum_{i=1}^m\alpha_i h_{\sigma}(x_i,\cdot) | x_i \in \mathbb{R}^n,\;\sum_{i=1}^m\alpha_i=1\textrm{, and }\alpha_i \geq 0\}$, where $\kappa_{\sigma}$ is a square integrable ``Parzen'' type of kernel, that is $h_{\sigma}$ is symmetric, nonnegative, has bounded integral (can be normalized), and shift invariant with $\sigma$ as scale parameter; $\mathcal{V}$ also defines an RKHS $\mathcal{K}$ on $\mathcal{G}$. Note that for any $g \in \mathcal{G}$, we have $\|\mathcal{V}(g,\cdot)\|_{\mathcal{K}} = \sum_{i,j}\alpha_i\alpha_j\mathcal{V}(x_i,x_j) \leq \mathcal{V}(x,x) < \infty $; therefore, $\mathcal{K}$ is a space of functionals on a bounded, albeit non-compact set. Notice that the cross information potential, by definition, is a positive definite function that is data dependent, and thus different from the instance-based kernel representation in machine learning. Nevertheless, the empirical estimator \eqref{eq:second_renyi_entropy} links both Hilbert space representations. 
If we construct the Gram matrix $\mathbf{K}$ with elements $K_{ij} = \kappa_{2\sigma}(x_i,x_j)$, it can be verified that \eqref{eq:second_renyi_entropy} corresponds to:
\begin{equation}\label{eq:trace_second_renyi_entropy}
\hat{H}_2(X) = -\log{\left(\frac{1}{n^2}\mathrm{tr}{\left(\mathbf{K}\mathbf{K}\right)}\right)} + C(\sigma).
\end{equation}
Here, $C(\sigma)$ takes care of the normalization factor that makes the Parzen window integrate to one. As we can see, the information potential estimator can be related to the Frobenius norm of the Gram matrix $\mathbf{K}$ defined as $\Vert\mathbf{K}\Vert^2 = \mathrm{tr}{\left(\mathbf{K}\mathbf{K}\right)}$. This observation is important because it motivates a generalization of the information potential based on the Gram matrices. In the following section, we extend this concept to other spectral norms and show how the properties of Renyi's definition of entropy carry over this generalization. The argument of the $\log$ function in \eqref{eq:Renyi_entropy} called the information potential has been studied in information theoretic learning and can be utilized interchangeably when our main goal is to maximize or minimize entropy since both quantities, entropy and information potential, are related by a strictly monotonic function. It is also worth mentioning that other relations between information theoretic quantities and positive definite kernels have also been previously established in the context of kernel methods, however, these have been derived from a different perspective and are meant to serve a different purpose. Mainly, these works aim at defining a Hilbert space embeddings for structured objects. Each object is represented by a probability model that is subsequently taken as the argument of a positive definite kernel defined on a space of probability measures. Examples of this approach are probability product kernels \cite{RKondor04}, Hilbert space embeddings of measures \cite{OBousquet04}, and the investigation of non-extensive information theoretic kernels on measures \cite{AMartins09}. We would like to remind the reader that our goal and scope are very different. We want to define an entropy measure directly from data, for which the theory of positive definite kernels arises during its construction. Defining a positive definite kernel between probability measures should not be confused with our goal.

%--------------------------------------------------------------------------------------------------------------------------------------
\section{Renyi's Entropy Axioms on Positive Definite Matrices}
%--------------------------------------------------------------------------------------------------------------------------------------
Renyi provided a set of axioms a function must fulfil to be considered a measure of information or entropy \cite{ARenyi60}. These axioms are presented in Appendix \ref{app:Renyi_Axioms} for the sake of completeness. Here, we provide a set of axioms for a matrix formulation of a measure of entropy. In particular, we employ nonnegative definite matrices, which can be considered a generalizations of nonnegative real numbers. 
%It is possible to define a partial ordering on this set by using positive definite matrices; for two Hermitian matrices $A,B \in M_n$, we say $A \succcurlyeq B$ if $A - B$ is positive definite. Likewise, $A \succ B$ means that $A - B$ is strictly positive definite.  

Our extension of Renyi's entropy to positive definite matrices uses the spectral theorem (see Appendix \ref{app:spectral_theorem} for details) to define matrix functions from scalar continuous functions. Let $A$ be a positive definite matrix with spectrum $\sigma(A)$, and $f(x)$ be a continuous real function defined for all $x \in \sigma(A) \subseteq \mathbb{R}$. The matrix function $f(A)$ is defined as $\sum_{i \in \mathcal{I}}f(\lambda_i)\mathbf{u}_i\mathbf{u}_i^{\mathrm{T}}$, where $\{\lambda_i\} \in \sigma(A)$ are the eigenvalues of $A$ and $\{\mathbf{u}_i\}$ the corresponding eigenvectors.

Consider the set $\Delta_n^{+}$ of positive definite matrices in the set of all real valued matrices of size $ n\times n$ denoted by $M_n$, for which $\mathrm{tr}{(A)} \leq 1$. This set is closed under finite convex combinations.
\begin{prop}\label{prop:matrix_entropy} Let $A \in \Delta_n^{+}$ and $B \in \Delta_n^{+}$ and also $\mathrm{tr}{(A)} = \mathrm{tr}{(B)} = 1$. The functional
\begin{equation}\label{eq:renyi_matrix_entropy}
S_{\alpha}(A) = \frac{1}{1-\alpha}\log_{2}{\left(\mathrm{tr}{A^{\alpha}}\right)},
\end{equation}
satisfies the following set of conditions:
\begin{enumerate}[(i)]
\item \label{property:matrix_symmetry} $S_{\alpha}(PAP^{*}) = S_{\alpha}(A)$ for any orthonormal matrix $P \in M_n$  
\item \label{property:matrix_continuity} $S_{\alpha}(pA)$ is a continuous function for $0 < p \leq 1$.
\item \label{property:matrix_max} $S_{\alpha}(\frac{1}{n}I) = \log_2{n}$.
\item \label{property:matrix_joint} $S_{\alpha}(A \otimes B) = S_{\alpha}(A) + S_{\alpha}(B)$.
\item \label{property:matrix_generalized_mean_value} If $AB = BA = \mathbf{0}$; then for the strictly monotonic and continuous function $g(x) = 2^{(\alpha-1)x}$ for $\alpha \neq 1$ and $\alpha > 0$, we have that:
\begin{equation}\label{eq:matrix_entropy_mean_value}
\begin{split}
S_\alpha(tA+(1-t)B) = & g^{-1}\left( tg(S_\alpha(A))+\right. \\
                      &            + \left. (1-t)g(S_\alpha(B))\right).
\end{split}
\end{equation} 
\end{enumerate}
\begin{prf} The proof of (\ref{property:matrix_symmetry}) follows from Theorem \ref{thm:functional_calculus_psd_matrix}. Take $A = U \Lambda U^*$, because $PU$ is also a unitary matrix we have that $f(A)$ = $f(PAP^*)$ (the trace functional is invariant under unitary transformations). For (\ref{property:matrix_continuity}), the proof reduces to the continuity of $\frac{1}{1-\alpha}\log_2(p)^\alpha$. For  (\ref{property:matrix_max}), a simple calculation yields $\mathrm{tr}{A^{\alpha}} = \left(\frac{1}{n}\right)^{\alpha-1}$. Now,  for property (\ref{property:matrix_joint}), notice that if $\mathrm{tr}{A} = \mathrm{tr}{B} = 1$, then,  $\mathrm{tr}{(A \otimes B)} = 1$. Since $A = U \Lambda U^*$ and $B = V \Gamma V^*$ we can write $A \otimes B = (U \otimes V)(\Lambda \otimes \Gamma)(U \otimes V)^*$, from which $\mathrm{tr}{(A \otimes B)^{\alpha}} = \mathrm{tr}{(\Lambda \otimes \Gamma)^{\alpha}} = \mathrm{tr}{(\Lambda^{\alpha})}\mathrm{tr}{(\Gamma^{\alpha})}$ and thus (\ref{property:matrix_joint}) is proved. Finally, (\ref{property:matrix_generalized_mean_value}) notice that for any integer power $k$ of $tA+(1-t)B$ we have: $(tA+(1-t)B)^k = (tA)^k + ((1-t)B)^k$ since $AB = BA = \mathbf{0}$. Under extra conditions such as $f(0) = 0$ the argument in the proof of Theorem \ref{thm:functional_calculus_psd_matrix} can be extended to this case. Since the eigen-spaces for the non-null eigenvalues of $A$ and $B$ are orthogonal we can simultaneously diagonalize $A$ and $B$ with the orthonormal matrix $U$, that is $A = U\Lambda U^*$ and $B = U\Gamma U^*$ where $\Lambda$ and $\Gamma$ are diagonal matrices containing the eigenvalues of $A$ and $B$ respectively. Since $AB = BA = \mathbf{0}$, then $\Lambda \Gamma = \mathbf{0}$. Under the extra condition $f(0) = 0$, we have that $f(tA+(1-t)B) = f(tA) + f((1-t)B)$, which yields the desired result for (\ref{property:matrix_generalized_mean_value}).\\
$\square$ 
\end{prf}
\end{prop}

Notice also that if $\mathrm{rank}(A) = 1$, $S_{\alpha} = 0$ for $\alpha \neq 0$. It it also possible to show that the identity matrix provides an upper bound for \eqref{eq:renyi_matrix_entropy}.
\begin{prop}\label{prop:information_inequality} Let $A \in \Delta_n^{+}$, and $\mathrm{tr}{(A)} = 1$. For $\alpha > 1$
\begin{equation}\label{eq:information_inequality}
S_{\alpha}(A) \leq S_{\alpha}(\frac{1}{n}I)
\end{equation}
\begin{prf} Let $\{\lambda_i\}$ be the set of eigenvalues of $A$. Then we have that, 
\begin{eqnarray}
S_{\alpha}(A)-S_{\alpha}(\frac{1}{n}I) & = & \frac{1}{1-\alpha}\log_2{\left[\frac{\mathrm{tr}{(AA^{\alpha-1})}}{n^{-(\alpha-1)}}\right]}; \\
~ & = & \frac{1}{1-\alpha}\log_2{\left[\mathrm{tr}{\left( A(nA)^{\alpha-1}\right)}\right]} ;\\
~ & = & \frac{1}{1-\alpha}\log_2{\left[\sum\limits_i \lambda_i f_\alpha(n\lambda_i)\right]} ;\\
\label{eq:information_inequality_Jensen_1} ~ & \leq & \frac{1}{1-\alpha}\sum\limits_i \lambda_i \log_2{f_\alpha(n\lambda_i)} ;\\
~ & = & -\sum\limits_i \lambda_i \log_2{\frac{\lambda_i}{\frac{1}{n}}};\\
\label{eq:information_inequality_Jensen_2} ~ & \leq & \log_2{\left[\sum\limits_i \lambda_i \frac{\frac{1}{n}}{\lambda_i}\right]}   = 0.
\end{eqnarray}
Where \eqref{eq:information_inequality_Jensen_1} and \eqref{eq:information_inequality_Jensen_2} are obtained from Jensen's inequality.\\ $\square$
\end{prf}
\end{prop}
The matrix functional in \eqref{eq:renyi_matrix_entropy} bears a lot of resemblance with well-known operational quantities from quantum information theory \cite{MOhya}, where the density matrix (operator) $\rho$ can be employed to compute expectation over an observable represented by the operator $A$ as $
\langle A \rangle = \mathrm{tr}\left(\rho A\right)$. For instance, Von Neumann's entropy \cite{VonNeumann} corresponds to 
\begin{equation}\label{eq:VonNeumann_entropy}
	S(\rho) = -\mathrm{tr}\left(\rho \log \rho \right),
\end{equation}
and quantum extensions of Renyi's entropy \cite{MMosonyi11} are given by
\begin{equation}\label{eq:Renyi_quantum_entropy}
	S_\alpha(\rho) = \frac{1}{1-\alpha}\log{\left[\mathrm{tr}\left(\rho^\alpha\right)\right]}.
\end{equation}
While some of the properties of \eqref{eq:VonNeumann_entropy} and \eqref{eq:Renyi_quantum_entropy} also apply to \eqref{eq:trace_second_renyi_entropy},  we need to point out that our approach to this functional is very different since we deal with the Gram matrices obtained from pairwise evaluations of a positive definite kernel on a data sample. Consequently, our analysis not only involves the functional but also the kernels employed to construct positive definite matrix. For instance rank one matrices that are normalized to have unit trace will have zero entropy. This means that the implicit dimensionality of the mapping induced by the positive definite kernel plays a key role. In the following, we extend the above matrix based entropy to a definition of joint entropy. This is achieved through the use of Hadamard products. 

\section{A Definition of Joint Entropy using Hadamard Products}
%--------------------------------------------------------------------------------------------------------------------------------------
Hadamard products $(A \circ B)_{ij} = A_{ij}B_{ij} $ are considered a rather simple form  of matrix products. However, the fact that the set of positive definite matrices is closed under this product motivates their study. Here, we use this property to extend our matrix-based definition of entropy to convey a joint representation of two random variables $X$ and $Y$. Given a sample $\{z_i = (x_i,y_i)\}_{i=1}^n $ of $n$ pairs representing two different types of measurements $x \in  \mathcal{X}$ and $y \in \mathcal{Y}$ obtained from the same realization, and the positive definite kernels $\kappa_1: \mathcal{X} \times \mathcal{X} \mapsto \mathbb{R}$ and $ \kappa_2: \mathcal{Y} \times \mathcal{Y} \mapsto \mathbb{R}$, we can form the matrices $A$ and $B$ by $A_{ij} = \kappa_1(x_i,x_j)$ and $B_{ij} = \kappa_2(y_i, y_j)$ and their Hadamard product $A \circ B$. The joint entropy is thus defined as:
\begin{equation}\label{eq:matrix_joint_entropy}
S_{\alpha}(A,B) = S_{\alpha}\left(\frac{A \circ B}{\mathrm{tr}(A \circ B)}\right)
\end{equation}
%In the above propositions \ref{prop:matrix_entropy} and \ref{prop:information_inequality}, we did not considered Hadamard products of positive definite matrices. Let $A$ and $B$ in $\Delta_n$ be matrices with unit trace for which there exists a relation between the elements $A_{ij}$ and $B_{ij}$ for all $i$ and $j$. The Hadamard product can be useful in developing analogues to joint entropies, where each one the matrices involved in the Hadamard product represents a random variable.
We can interpret the Hadamard product as computing a product kernel $\kappa((x_i,y_i),(x_j,y_j)) = \kappa_1(x_i,x_j)\kappa_2(y_i,y_j)$, which can be interpreted as computing the measure of entropy of a random element defined by the pair $(X,Y)$. Nevertheless, as we will show in this section, for this notion of joint entropy to be compatible with the individual entropies of each one of the components, we need to impose some extra conditions on the positive definite matrices. For example, we should expect the joint entropy to be always larger than the individual entropies of its components. Before we present the main result of the section, we need to introduce the concept of majorization and some results pertaining the ordering that arises from it.
\begin{defi}\emph{(Majorization): }Let $p$ and $q$ be two nonnegative vectors in $\mathbb{R}^n$ such that $\sum_{i=1}^n p_i = \sum_{i=1}^n q_i$. We say $p \preccurlyeq q$, $q$ majorizes $p$, if their respective ordered sequences $p_{[1]} \geq p_{[2]} \geq \dots \geq p_{[n]}$  and $q_{[1]} \geq q_{[2]} \geq \dots \geq q_{[n]}$ denoted by $\{p_{[i]}\}_{i=1}^n$ and $\{p_{[i]}\}_{i=1}^n$, satisfy: 
\begin{equation}
\sum_{i=1}^k p_{[i]}  \leq \sum_{i=1}^k q_{[i]}\:\: \textrm{for} \:\: k =1,\dots, n
\end{equation}
\end{defi}
It can be shown that if $p \preccurlyeq q$ then $p = Aq$ for some doubly stochastic matrix $A$ \cite{RBhatia}. It is also easy to verify that if $p \preccurlyeq q$ and $p \preccurlyeq h$ then $p \preccurlyeq tq +  (1-t)h$ for $t \in [0,1]$. The majorization order is important because it can be associated with the definition of \emph{Schur-concave (convex)} functions. A real valued function $f$ on $\mathbb{R}^n$ is called Schur-convex if $p \preccurlyeq q$ implies $f(p) \leq f(q)$ and Schur-concave if $f(q) \leq f(p)$. 
\begin{lmm}\label{lmm:Renyi_schur_concavity} The function $f_{\alpha}:\mathcal{S}^n \mapsto \mathbb{R}_{+}$ ($\mathcal{S}^n$ denotes the $n$ dimensional simplex), defined as,
\begin{equation}
f_{\alpha}(p) = \frac{1}{1-\alpha}\log_2{\sum\limits_{i=1}^n p_i^{\alpha}}, 
\end{equation}
is Schur-concave for $\alpha>0$.
\end{lmm}
Notice that, Schur-concavity (Schur-convexity) ought not be confused with the more common definition of concavity (convexity) of a function. We are now ready to state the inequality for Hadamard products that makes the definition of joint entropy compatible with the individual entropies of its components.  
\begin{prop}\label{prop:hadamard_inequalities} Let $A$ and $B$ be two $n \times n$ positive definite matrices with trace $1$ with nonnegative entries, and $A_{ii} = \frac{1}{n}$ for $i=1,2,\dots,n$. Then, the following inequalities hold:
\begin{enumerate}[(i)]
\item \begin{equation}\label{eq:hadamard_conditional_entropy_inequality} 
 S_{\alpha}\left(\frac{A \circ B}{\mathrm{tr}(A \circ B)}\right) \geq S_{\alpha}(B),
\end{equation}
\item and \begin{equation}\label{eq:hadamard_joint_entropy_inequality} 
 S_{\alpha}\left(\frac{A \circ B}{\mathrm{tr}(A \circ B)}\right) \leq S_{\alpha}(A)+S_{\alpha}(B).
\end{equation}
\end{enumerate}
\begin{prf} In proving \eqref{eq:hadamard_conditional_entropy_inequality} and \eqref{eq:hadamard_joint_entropy_inequality}, we will use the fact that $S_{\alpha}$ preserves the (inverse) majorization order of nonnegative sequences on the $n$-dimensional simplex.
First consider the identity
\[
x^{\mathrm{T}}(A \circ B)x = \mathrm{tr}{\left(A D_x B D_x\right)}
\]
In particular, if $\{x_i\}_{i=1}^n$ is an orthonormal basis for $\mathbb{R}^n$,
\[
\mathrm{tr}{\left(A \circ B \right)} = \sum\limits_{i=1}^nx_i^{\mathrm{T}}(A \circ B)x_i 
\]
If we let  $\{x_i\}_{i=1}^n$ be the eigenvectors of $A \circ B$ ordered according to their respective eigenvalues in decreasing order, then,
\begin{eqnarray}
\nonumber \sum\limits_{i=1}^k x_i^{\mathrm{T}}(A \circ B)x_i & = & \sum\limits_{i=1}^k \mathrm{tr}\left(A D_{x_i} B D_{x_i}\right) \\
\nonumber & \leq & \frac{1}{n}\sum\limits_{i=1}^k \mathrm{tr}\left(\mathbf{1}\mathbf{1}^{\mathrm{T}} D_{x_i} B D_{x_i}\right) \\
\nonumber & = & \frac{1}{n}\sum\limits_{i=1}^k x_i^{\mathrm{T}} B x_i \\
\label{eq:majorization_conditional_inequality} & \leq & \frac{1}{n}\sum\limits_{i=1}^k y_i^{\mathrm{T}} B y_i,
\end{eqnarray}
where $k=1,\dots,n$ and $\{y_i\}_{i=1}^n$ are the eigenvectors of $B$ ordered according to their respective eigenvalues in decreasing order. The inequality \eqref{eq:majorization_conditional_inequality} is equivalent  to $n \lambda(A \circ B) \preccurlyeq \lambda(B)$, that is, the sequence of eigenvalues of $(A \circ B)/\mathrm{tr}{(A \circ B)}$ is majorized by the sequence of eigenvalues of $B$, which implies \eqref{eq:hadamard_conditional_entropy_inequality}.\\
To prove \eqref{eq:hadamard_joint_entropy_inequality} notice that for $A$ we have two extreme cases $A = \frac{1}{n}I$ and $A=\frac{1}{n} \mathbf{1}\mathbf{1}^{\mathrm{T}}$. Taking $A=\frac{1}{n} \mathbf{1}\mathbf{1}^{\mathrm{T}}$  we have that 
\begin{equation}\label{eq:upper_joint_entropy_extreme}
\sum\limits_{i=1}^k \lambda_i(B) = n \sum\limits_{i=1}^k \frac{1}{n} \mathrm{tr}{\left( \mathbf{1}\mathbf{1}^{\mathrm{T}} D_{x_i} B D_{x_i} \right)} = \sum\limits_{i=1}^k \lambda_i \left( \frac{A\circ B}{\mathrm{tr}(A\circ B)} \right)
\end{equation}
the other extreme case where $A = \frac{1}{n} I$ we have,
\begin{equation}\label{eq:lower_joint_entropy_extreme}
\frac{1}{n}\sum\limits_{i=1}^k \lambda_i(B) \leq \frac{1}{n} \leq n \sum\limits_{i=1}^k \frac{1}{n}d_i(B) = \sum\limits_{i=1}^k \lambda_i\left(\frac{A\circ B}{\mathrm{tr}(A \circ B)}\right)
\end{equation}
where $\{\lambda_i(X)\}$ are the eigenvalues of $X$ in decreasing order and $\{d_i(X)\}$ are the elements of the diagonal of $X$ ordered in decreasing order. The inequalities \eqref{eq:upper_joint_entropy_extreme} and \eqref{eq:lower_joint_entropy_extreme} imply \eqref{eq:hadamard_joint_entropy_inequality}.\\
$\square$
\end{prf}
\end{prop}
\subsection{Conditional Entropy}
In Shannon's definition the conditional entropy of $X$ given $Y$, that can be calculated as the difference $H(X|Y) = H(X,Y)-H(Y)$, is understood as the uncertainty about $X$ that remains after observing $Y$ given that the joint distribution of $X$ and $Y$ is known. In contrast, Renyi's conditional entropy allows multiple definitions and there is no general consensus on which definition should be adopted \cite{ATeixeira12}. In our work, based on \eqref{eq:hadamard_conditional_entropy_inequality}, we have chosen the following definition:
\begin{equation}\label{eq:conditonal_entropy_gap}
S_{\alpha}(A|B) =  S_{\alpha}\left(\frac{A \circ B}{\mathrm{tr}{(A\circ B)}}\right) - S_{\alpha}(B)
\end{equation}
for positive semidefinite $A$ and $B$ with nonnegative entries and unit trace, such that $A_{ii} = \frac{1}{n}$ for all $i=1,\dots,n$. Our definition relies on the inequalities \eqref{eq:hadamard_conditional_entropy_inequality} and \eqref{eq:hadamard_joint_entropy_inequality}, which make the above quantity \eqref{eq:conditonal_entropy_gap} nonnegative and upper bounded by $S_{\alpha}(A)$. 
\subsection{Mutual Information}
The mutual information of a pair of random variables $X$ and $Y$ is associated with the reduction of the uncertainty from having only knowledge about their marginal distributions to full knowledge their joint distribution. In the Shannon definition this information gain can be expressed as:
\begin{equation}
I(X;Y) = H(X) + H(Y) - H(X,Y)
\end{equation}
where $H(X)$ and $H(Y)$ are the marginal entropies of $X$ and $Y$, and $H(X,Y)$ is their joint entropy. In analogy, we can compute the quantity:
\begin{equation}\label{eq:tensor_entropy_gap}
I_{\alpha}(A;B) = S_{\alpha}(A) + S_{\alpha}(B) - S_{\alpha}\left(\frac{A \circ B}{\mathrm{tr}{(A\circ B)}}\right)
\end{equation}
for positive semidefinite $A$ and $B$ with nonnegative entries and unit trace, such that $A_{ii}  = B_{ii}= \frac{1}{n}$ for all $i=1,\dots,n$. Notice that the above quantity is nonnegative and satisfies
\[
S_{\alpha}(A) \geq I_{\alpha}(A;A).
\]
Here, we want to emphasize that this extension relies on proposition \ref{prop:hadamard_inequalities} and differs from the common definition of Renyi's mutual information, which is based on Renyi's $\alpha$-order divergence between the joint distribution and the product of the marginal distributions. 
As we can see, for the above inequalities to hold, we need extra conditions on the elements of $A$ and $B$. First, notice that the elements of the diagonal must be $\frac{1}{n}$, which implies that $\kappa_i(x,x) > 0$ for all $x \in \mathcal{X}$. A matrix with equal elements can be obtained by a simple normalization
\begin{equation}\label{eq:normalization}
\hat{A}_{ij} = \frac{A_{ij}}{\sqrt{A_{ii}}\sqrt{A_{jj}}}.
\end{equation}
Now, let us focus on the Hadamard product itself. Let $A^{\circ r}$ denote the entry-wise $r$th power of $A$, that is, $(A^{\circ r})_{ij} = A_{ij}^r$, where $r \in \mathbb{R}^+$. In this case we also require all the entries of $A$ to be nonnegative. For the set of normalized diagonal matrices, the product $A^{\circ r}\circ B^{\circ (1-r)}$  with $r \in (0,1)$ preserves the normalization condition. Nonetheless, the positive definiteness of this product is only guaranteed for a special class of matrices known as infinitely divisible. In the following section, we will see how infinitely divisible matrices relate Hadamard products with the concatenation of representations of the variables we want to analyze jointly. 
\section{Information Theoretic Functionals based on Infinitely Divisible Matrices }

%Let us recall \eqref{eq:hadamard_entropy_inequality}, $S_{\alpha}(A \otimes B) \geq S_{\alpha}\left((A \circ B)/\mathrm{tr}(A \circ B)\right)$.
%
%Infinitely divisible matrices, can exhibit the following properties:
%\begin{itemize}
%\item Positive definite by definition. 
%\item Non-negative entries.
%\item In connection to the example about class encoding vectors, we can note that $\mathbf{L} = \frac{1}{n}\mathbf{ll}^{\mathrm{T}}$ is not infinitely divisible, whereas $\mathbf{L} = \mathbf{MM}^{\mathrm{T}}$ does. 
%\end{itemize}
According to the Schur product theorem $A \succcurlyeq 0$ implies that $A^{\circ n} = A \circ A \circ \cdots \circ A \succcurlyeq 0$ for any positive integer $n$. However, the fractional powers of $A \succcurlyeq 0$, denoted $A^{\circ \frac{1}{m}} $ do not need to be positive definite for any positive integer $m$. This is only true for a special class of matrices called infinitely divisible \cite{RBhatia06,RHorn69}.
\begin{defi}Suppose that $A \succcurlyeq 0$ and $a_{ij} \geq 0$ for all $i$ and $j$. $A$ is said to be infinitely divisible if $A^{\circ r} \succcurlyeq 0$ for every nonnegative $r$.
\end{defi}

Infinitely divisibility and negative definiteness are intimately related as we can see from the following proposition \cite{CBerg}
\begin{prop}\label{prop:infinitely_divisibility_negative_definiteness} If $A$ is infinitely divisible, then the matrix $B_{ij} =-\log{A_{ij}}$ is negative definite. 
\end{prop} 
Therefore, infinitely divisible matrices can be linked to isometric embeddings into Hilbert spaces \cite{ISchoenberg38}. If we construct the matrix,
\begin{equation}\label{eq:Negative_definite_Hilbertian_metric}
D_{ij} = B_{ij}-\frac{1}{2}(B_{ii}+B_{jj}), 
\end{equation}
using the matrix $\mathbf{B}$ from proposition \ref{prop:infinitely_divisibility_negative_definiteness}, there exist a Hilbert space $\mathcal{H}$ and a mapping $\phi$ such that
\begin{equation}\label{eq:Hilbertian_metric}
D_{ij} = \Vert\phi(i) - \phi(j) \Vert_{\mathcal{H}}^2.
\end{equation}
Moreover, notice that if $A$ is positive definite $-A$ is negative definite  and $\exp{A_{ij}}$ is infinitely divisible \cite{CBerg}. In a similar way, we can construct a matrix,
\begin{equation}\label{eq:Positive_definite_Hilbertian_metric}
D_{ij} = -A_{ij}+\frac{1}{2}(A_{ii}+A_{jj}), 
\end{equation}
with the same property \eqref{eq:Hilbertian_metric}. This relation between \eqref{eq:Negative_definite_Hilbertian_metric} and \eqref{eq:Positive_definite_Hilbertian_metric} suggests a normalization for infinitely divisible matrices with non-zero diagonal elements that can be formalized in the following theorem.

\begin{thm}\label{thm:normalization_theorem} Let $\mathcal{X}$ be a nonempty set and $d_1$ and $d_2$ two semi-metrics defined on it, such that for any set $\{x_i\}_{i=1}^n$,
\begin{equation}\label{eq:condtionally_negative_definite_condition}
\sum\limits_{i,j = 1}^n\alpha_i\alpha_j d_{\ell}^2(x_i,x_j)  \leq 0,\: \textrm{and}\: \ell=1,2  
\end{equation}
for any $\bm{\alpha} \in \mathbb{R}^{n}$, and $\sum_{i=1}^n\alpha_i = 0$. Consider the matrices $A_{ij}^{(\ell)} = \exp{-d_{\ell}^2(x_i,x_j)}$ and their normalizations $\hat{A}^{(\ell)}$, defined by:
\begin{equation}\label{eq:infinitely_divisible_normalization}
\hat{A}_{ij}^{(\ell)} = \frac{A_{ij}^{(\ell)}}{\sqrt{A_{ii}^{(\ell)}}\sqrt{A_{jj}^ {(\ell)}}}.
\end{equation}
Then, if $\hat{A}^{(1)} = \hat{A}^{(2)}$ for any finite set $\{x_i\}_{i=1}^n \subseteq \mathcal{X}$, there exist isometrically isomorphic Hilbert spaces $\mathcal{H}_1$ and $\mathcal{H}_2$, that contain the Hilbert space embeddings of the metric spaces $(\mathcal{X},d_{\ell})$, $\ell=1,2$. Moreover, $\hat{A}^{(\ell)}$ are infinitely  divisible. 
%\begin{prf}First, 
%\end{prf}
\end{thm}  
Notice that the  normalization procedure for infinitely divisible matrices proposed in Theorem \ref{thm:normalization_theorem} corresponds to the matrix with maximum entropy among all matrices for which the Hilbert space embeddings are isometrically isomorphic (See \eqref{eq:upper_joint_entropy_extreme}).

\begin{figure}
\centering
\includegraphics[width =8.5cm]{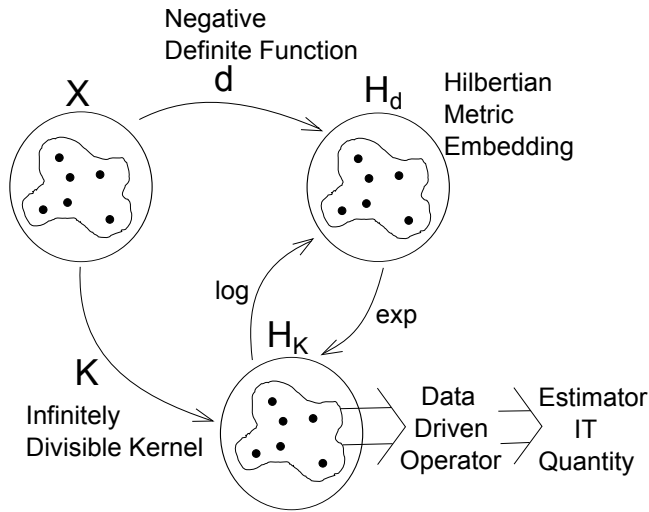}
\caption{Spaces involved in the infinitely divisible matrix framework}\label{fig:spaces} 
\end{figure}

The link between infinitely divisible kernels and negative definite functions is depicted in Figure \ref{fig:spaces}. To obtain a the Gram matrix employed by the entropy functional there is a direct path, provided that an infinitely divisible kernel $\kappa$ on $\mathcal{X}$ that can be normalized based on Theorem \ref{thm:normalization_theorem} is at hand. The second alternative path uses a negative definite function $d$ which can be given or defined from any positive definite kernel on $\mathcal{X}$. The link between the spaces $\mathcal{H}_\kappa$  and $\mathcal{H}_d$ is established by using the $\log$ and $\exp$ functions, accordingly. As we will see in the following section, the matrix based entropy is an estimator of a quantity that can be interpreted as a functional on an operator embedding of the probability measure from which data is drawn. 
%--------------------------------------------------------------------------------------------------------------------------------------
\section{Gram Matrices and Operator Embeddings of Probability Measures}
%--------------------------------------------------------------------------------------------------------------------------------------
So far, we have focused the attention on the properties of the matrix based entropies for a given Gram matrix. In this section, we will look at the relation between the Gram matrix and certain operators acting on the elements of the RKHS induced by a kernel $\kappa$. In this sense, we can understand these operators as image points of an embedding of probability distributions. This will enable us to study the converge properties of the matrix based entropies as estimators of well-defined population entropies. Let $\left( \mathcal{X},\mathcal{B}_{\mathcal{X}},P_{\mathcal{X}}\right)$  be a countably generated measure space. Let $\kappa : \mathcal{X} \times \mathcal{X} \mapsto \mathbb{R}$ be a reproducing kernel and the mapping $\phi: \mathcal{X} \mapsto \mathcal{H} $ such that $\kappa(x,y) = \langle \phi(x),\phi(y) \rangle$, and\footnote{Any nonnegative normalized kernel satisfies this assumption as long as $\phi(x) \neq 0$ for all $x \in \mathcal{X}$}:
\begin{equation}\label{eq:normalized_kernel_condition}
\begin{split}
\EV_{X}\left[ \kappa(X,X) \right] & = \EV_{X}\left[ \Vert\phi(X) \Vert^2 \right] \\ & = \int\limits_{\mathcal{X}}\langle \phi(x),\phi(x) \rangle \mathrm{d}P_{\mathcal{X}}(x) = 1
\end{split}
\end{equation}
Since $\EV_{X}\left[ \Vert\phi(X) \Vert^2 \right] < \infty$ we can define an operator $G : \mathcal{H} \mapsto \mathcal{H}$ through the following bilinear form\footnote{Notice, that $f \in \mathcal{H} \Rightarrow f \in L_{2}(P\mathcal{X})$. First, $\vert f(x) \vert = \vert\langle f,\phi(x) \rangle \vert  \leq \Vert f \Vert \kappa(x,x)^{\frac{1}{2}}$, and thus $f(x)^2 \leq \Vert f \Vert^2 \kappa(x,x)$. Since $\int\kappa(x,x)\mathrm{d}P_{\mathcal{X}} = 1$, we have $\Vert f \Vert_{2}^2 = \int f^2 \mathrm{d}P_{\mathcal{X}} \leq \Vert f \Vert^2$}:
\begin{equation}\label{eq:operator_bilinear_form}
\mathcal{G}(f,g) = \langle f, Gg \rangle = \int\limits_{\mathcal{X}}\langle f,\phi(x) \rangle \langle \phi(x),g \rangle \mathrm{d}P_{\mathcal{X}}(x)
\end{equation}
notice that $f$ and $g$ belong to $\mathcal{H}$ and from the reproducing property of $\kappa$, we have that $f(x) = \langle f,\phi(x) \rangle$ and thus $\mathcal{G}(f,g) = \EV_{X}\left[ f(X)g(X) \right]$. From the normalization condition \eqref{eq:normalized_kernel_condition} we have that:
\begin{equation}\label{eq:trace_operator}
\begin{split}
\mathrm{tr}(G) & = \sum\limits_{i=1}^{N_{\mathcal{H}}}\mathcal{G}(\psi_i,\psi_i) \\  & = \sum\limits_{i=1}^{N_{\mathcal{H}}}\int\limits_{\mathcal{X}}\langle \psi_i,\phi(x) \rangle \langle \phi(x),\psi_i \rangle \mathrm{d}P_{\mathcal{X}}(x) = 1
\end{split}
\end{equation}
where $\left\{ \psi_i \right\} _{i=1}^{N_{\mathcal{H}}}$ is a complete orthonormal basis for $\mathcal{H}$. The operator $G$ is trace class and therefore it is also Hilbert-Schmidt and compact. 

\subsection{The trace of $G^{\alpha}$}
As we have seen, $G$ defines a bilinear form $\mathcal{G}$ that coincides with the correlation of functions on $\mathcal{X}$ that belong to the RKHS induced by $\kappa$. Let us look at the case $\alpha = 2$, which is the initial motivation of this study and has been extensively treated in information theoretic learning in relation to plug in estimators of Renyi's entropy \cite{JPrincipe}. This case is also important because of its links to maximum discrepancy and Hilbert Schmidt norms \cite{AGretton08}. Notice that,

\begin{eqnarray}\label{eq:Second_order_operator_definition}
\nonumber \mathrm{tr}\left(G^2\right) & = & \sum\limits_{i=1}^{N_\mathcal{H}}\langle \psi_i , G^2\psi_i \rangle = \sum\limits_{i=1}^{N_\mathcal{H}}\langle G\psi_i , G\psi_i \rangle \\
\nonumber & = & \sum\limits_{i=1}^{N_\mathcal{H}}\Vert G\psi_i \Vert^2 = \Vert G \Vert_{\textrm{HS}}^2 \\
\nonumber & = & \sum\limits_{i=1}^{N_\mathcal{H}}\int\limits_{\mathcal{X}}\int\limits_{\mathcal{X}}\langle\phi(x)\langle \phi(x),\psi_i\rangle , \dots \\
\nonumber & & \qquad \qquad  \phi(y)\langle \phi(y),\psi_i\rangle \rangle  \mathrm{d}P_{\mathcal{X}}(x) \mathrm{d}P_{\mathcal{X}}(y)\\
\nonumber & = & \int\limits_{\mathcal{X}}\int\limits_{\mathcal{X}}\langle\phi(x) , \phi(y) \rangle \langle \phi(x), \dots \\
\nonumber & & \qquad \qquad \sum\limits_{i=1}^{N_\mathcal{H}}\psi_i\langle \psi_i, \phi(y)\rangle\rangle \mathrm{d}P_{\mathcal{X}}(x) \mathrm{d}P_{\mathcal{X}}(y)\\
\nonumber & = & \int\limits_{\mathcal{X}}\int\limits_{\mathcal{X}}\langle\phi(x) , \phi(y) \rangle ^2 \mathrm{d}P_{\mathcal{X}}(x) \mathrm{d}P_{\mathcal{X}}(y)\\
 & = &\left\Vert\mu_X\right\Vert_{\mathcal{K}}^2,
\normalsize
\end{eqnarray}

where $\left\Vert\mu_X\right\Vert_{\mathcal{K}}^2$ denotes the squared norm of a the a mapping $P_{\mathcal{X}} \mapsto \mu_{X}$ in the RKHS $\mathcal{K}$ induced by the kernel $\kappa^2(x,y) = \kappa(x,y)\kappa(x,y)$. In the more general case of any $\alpha > 1$ we have,
\begin{eqnarray}\label{eq:any_order_operator_definition}
\nonumber \mathrm{tr}\left(G^\alpha \right) & = & \sum\limits_{i=1}^{N_\mathcal{H}}\langle \psi_i , G^{\alpha}\psi_i \rangle = \sum\limits_{i=1}^{N_\mathcal{H}}\langle G\psi_i , G^{\alpha-1}\psi_i \rangle \\
\nonumber & = & \sum\limits_{i=1}^{N_\mathcal{H}}\int\limits_{\mathcal{X}}\langle \psi_i, \phi(x)\rangle \langle \phi(x),G^{\alpha-1} \psi_i\rangle  \mathrm{d}P_{\mathcal{X}}(x)\\
\nonumber & = & \int\limits_{\mathcal{X}}\langle \phi(x) , G^{\alpha-1} \phi(x) \rangle \mathrm{d}P_{\mathcal{X}}(x) \\
\nonumber & = & \int\limits_{\mathcal{X}} h(x,x) \mathrm{d}P_{\mathcal{X}}(x). \\
\normalsize
\end{eqnarray}
Notice that $h(x,y)$ itself, is a positive definite function on $\mathcal{X} \times \mathcal{X}$ that also depends on $P_{\mathcal{X}}(x)$.\\  

\subsection{Estimating the Spectrum of $G$} 
By definition, the bilinear form $\mathcal{G}$ is a positive definite kernel in $\mathcal{H}$ since
\begin{equation}\label{eq:positive_definiteness_G}
\sum\limits_{i,j =1}^N \alpha_i\alpha_j\mathcal{G}(f_i,f_j) \geq 0
\end{equation}
for any finite set $\{f_i\}_{i=1}^N \subseteq \mathcal{H}$. Notice from \eqref{eq:operator_bilinear_form} $\mathcal{G}$ is symmetric and thus  $G$ is self adjoint. Moreover, since $\mathcal{G}$ is positive definite, it can be shown that $G$ is a positive definite operator. 
Instead of dealing directly with the spectrum of $G$, for which we should know the probability measure $P_{\mathcal{X}}$, we are going to look at the spectrum of $\widehat{G}_N$ and the convergence properties of this operator. Based on the empirical distribution $P_N = \frac{1}{N}\sum_{i=1}^N\delta_{x_i}(x)$, the empirical version $\widehat{G}_N$ of $G$ obtained from a sample $\{ x_i \}$ of size $N$ is given by:
\begin{equation}\label{eq:empirical_version_G}
\begin{split}
\langle f,\widehat{G}_Ng \rangle= \widehat{\mathcal{G}}_{N}(f,g) & = \int\limits_{\mathcal{X}}\langle f,\phi(x) \rangle \langle \phi(x),g \rangle \mathrm{d}P_N(x)\\
& = \frac{1}{N}\sum\limits_{i=1}^N\langle f,\phi(x_i) \rangle \langle \phi(x_i),g \rangle
\end{split}
\end{equation}
Note that $\EV[\widehat{G}_N] = G$. In the definition of the entropy like quantity for positive definite matrices, we employ functional calculus using the spectral theorem to compute $\mathrm{tr}(A^\alpha)$. In particular, we consider the Gram matrix $\mathbf{K}$ constructed by all pairwise evaluations of a normalized  infinitely divisible kernel $\kappa$ and scale it by $\frac{1}{N}$ such that $\frac{1}{N}\sum_{i=1}^N\kappa(x_i,x_i) = 1$. The above scaling can be thought as normalizing the kernel such that for the empirical distribution $P_N$,
\begin{equation}\label{eq:empirical_normalized_kernel_condition}
\begin{split}
\EV_{\textrm{emp}}\left[ \kappa(X,X) \right]&  = \EV_{\textrm{emp}}\left[ \Vert\phi(X) \Vert^2 \right] \\
&  = \int\limits_{\mathcal{X}}\langle \phi(x),\phi(x) \rangle \mathrm{d}P_{N}(x) \\
&  = \frac{1}{N}\sum\limits_{i=1}^N\kappa(x_i,x_i) = 1.
\end{split}
\end{equation}
The following proposition relates the spectrum of $\widehat{G}_N$ with the eigenvalues of the Gram matrix $\mathbf{K}$
\begin{prop}\label{prop:spectrum_empirical_G}\emph{(Spectrum of $\widehat{G}_N$): }For a sample $\{ x_i \}_{i=1}^N$, let $\widehat{G}_N$ be defined as in \eqref{eq:empirical_version_G}, and let $\mathbf{K}$ be the Gram matrix of products $K_{ij} = \langle \phi(x_i),\phi(x_j) \rangle $. Then,  $\widehat{G}_N$ has at most $N$ positive eigenvalues $\lambda_k$ satisfying:
\begin{equation}\label{eq:spectrum_empirical_G}
\frac{1}{N}\mathbf{K}\bm{\alpha}_i = \lambda_i\bm{\alpha}_i. 
\end{equation}  
Moreover, $N\lambda_i$ are all the positive eigenvalues of $\mathbf{K}$.
\begin{prf}
First notice that for all $f \perp \mathrm{span}\left\{\phi(x_i)\right\}$, we have $\widehat{G}_Nf = 0$, and thus any eigenvector with a corresponding positive eigenvalue must belong to the $\mathrm{span}\left\{\phi(x_i)\right\}$, which is an $N$ dimensional subspace and therefore, since $\widehat{G}_N$ is normal there can be at most $N$ positive eigenvalues. Now let $v$ be an eigenvector of $\widehat{G}_N$,, we have that 
\[
\langle \cdot,\widehat{G}_N v \rangle = \frac{1}{N}\sum\limits_{j=1}^N \langle\cdot,\phi(x_j) \rangle\langle \phi(x_j),v \rangle = \langle \cdot,\lambda v \rangle.
\] 
Then, for each $\phi(x_i)$ it is true that 
\[
\langle \phi(x_i),\widehat{G}_N v \rangle = \frac{1}{N}\sum\limits_{j=1}^N\langle \phi(x_i),\phi(x_j) \rangle\langle \phi(x_j),v \rangle = \lambda \langle \phi(x_i),v \rangle.
\]
By taking $\alpha_i = \langle \phi(x_i),v \rangle$ we can form the following system of equations:
\begin{equation}
\frac{1}{N}\mathbf{K}\bm{\alpha} = \lambda\bm{\alpha}. 
\end{equation} 
which is true for all positive eigenvalues of $\widehat{G}_N$.\\
$\square$
\end{prf}
\end{prop}
As a consequence of the relation established in Proposition \ref{prop:spectrum_empirical_G} the following is also true  
\begin{equation}\label{eq:empirical_trace_power_estimator}
\textrm{tr}\left[{\widehat{G}_N}^{\alpha}\right] = \textrm{tr}\left[\left(\frac{1}{N}\mathbf{K}\right)^{\alpha}\right]. 
\end{equation}

Now, we will focus the attention on the properties of \eqref{eq:empirical_trace_power_estimator} as an estimator of $\mathrm{tr}\left( G^{\alpha}\right)$. The first result that relates the spectrum of $G$ with the eigenvalues of $K$ has been previously considered in \cite{LRosasco10} in the context of learning algorithms that are based on estimating the eigenvalues and eigenfunctions of operators defined by a kernel. The following theorem found in \cite{TKato87} is a variational characterization of the discrete spectrum (eigenvalues) of a compact operator in a separable Hilbert space. 
\begin{thm}\label{thm:variational_characterization_eigenvalues}
Let $A$, $B$ be self adjoint operators in a separable Hilbert space $\mathcal{H}$, such that $B = A+C$, where $C$ is a compact selfadjoint operator. Let $\{\gamma_k\}$ be an enumeration of nonzero eigenvalues of $C$. Then there exists extended enumerations $\{\alpha_j\}$, $\{\beta_j\}$ of discrete eigenvalues for $A$, $B$, respectively, such that:
\begin{equation}
\sum\limits_{j}\varphi(\beta_j-\alpha_j) \leq   \sum\limits_{k}\varphi(\gamma_k),
\end{equation}
where $\varphi$ is any nonnegative convex function on $\mathbb{R}$, and $\varphi(0) = 0$.
\end{thm} 
The definition of extended enumeration $\{\alpha_i\}$ according to Theorem \ref{thm:variational_characterization_eigenvalues}   
means that for a selfadjoint operator $A$ in $\mathcal{H}$ only the discrete eigenvalues with finite multiplicity $m$ are listed $m$ times and any other values are listed as zero. 
If we have a bounded kernel, which in the case of a normalized version of the infinitely divisible matrix is always the case, we can apply Hoeffding's inequality. Let ${\Phi_i}$ be a sequence of zero mean, independent  random variables taking values in a separable Hilbert space such that $\Vert \Phi_i \Vert < C$ for all $i$ then:
\begin{equation}\label{eq:hoeffding_inequality}
\mathrm{Pr}\left[ \left\Vert\frac{1}{N}\sum\limits_{i=1}^{n}\Phi_n \right\Vert \geq \varepsilon \right] \leq 2 \exp{-\frac{N\varepsilon^2}{2C^2}}.
\end{equation} 
Note that $(\widehat{G}_N-G)$ is a Hilbert-Schmidt operator and $\EV[\widehat{G}_N] = G$. 
Combining \eqref{eq:hoeffding_inequality} with Theorem \ref{thm:variational_characterization_eigenvalues}, yields the following result.
\begin{thm}\label{thm:convergence_of_G} For a positive definite kernel $\kappa$ satisfying \eqref{eq:normalized_kernel_condition}, and $\kappa(x,x) \leq C$. Let ${\lambda_i}$ and ${\widehat{\lambda}_i}$ the extended enumerations of the discrete eigenvalues of $G$ and $\widehat{G}_{N}$, respectively. Then, with probability $1-\delta$
\begin{equation}\label{eq:bound_eignevalues}
\left(\sum\limits_{i}(\lambda_i-\widehat{\lambda}_i)^2\right)^{\frac{1}{2}} \leq C\sqrt{\frac{2\log{\frac{2}{\delta}}}{N}}
\end{equation}
\begin{prf}
Apply the result of Theorem \ref{thm:variational_characterization_eigenvalues} using $\varphi(x) = x^2$. \\
$\square$
\end{prf}
\end{thm}
As a consequence of the convergence of the spectrum of $\widehat{G}_N$ to the spectrum of $G$ and the continuity of $\mathrm{tr}\left( A^\alpha\right)$, the difference between $\mathrm{tr}\left(G^\alpha \right)$ and $\mathrm{tr}\left(\widehat{G}_N^\alpha \right)$ can be bounded as well. Under the conditions of Theorem \ref{thm:convergence_of_G} and for $\alpha > 1$, with probability $1-\delta$, 
\begin{equation}\label{eq:bound_traces}
\left\vert \mathrm{tr}\left(G^\alpha \right) - \mathrm{tr}\left(\widehat{G}_N^\alpha \right) \right\vert \leq \alpha C\sqrt{\frac{2\log{\frac{2}{\delta}}}{N}}. 
\end{equation}
The constant $\alpha$ in the bounding term \eqref{eq:bound_traces} results from raising the eigenvalues of the Gram matrix to the power of $\alpha$. Although it seems that as $\alpha$ increases a larger number of points is required for accurate estimation, in practice, the bound can be much lower if the dominant eigenvalues fall below $\alpha^{-\frac{1}{\alpha - 1}}$. In this case the constant can be set to be $1$.\\
Following a similar approach to \cite{CCortes12}, we can extend the convergence results of the matrix based entropy to our definition of conditional entropy. Let us consider the argument of the $\log$ function in our definition of conditional entropy $S_{\alpha}(B|A)$, 
\begin{equation}\label{eq:conditional_entropy_ratio}
\frac{\mathrm{tr}\left[A^\alpha\right]}{\mathrm{tr}\left[\left(\frac{A \circ B}{\mathrm{tr}{(A\circ B)}}\right)^\alpha\right]}.
\end{equation}
Let $\kappa_{\mathcal{X}} : \mathcal{X} \times \mathcal{X} \mapsto \mathbb{R}$ and $\kappa_{\mathcal{Y}} : \mathcal{Y} \times \mathcal{Y} \mapsto \mathbb{R}$ be reproducing kernels and the mappings $\phi: \mathcal{X} \mapsto \mathcal{H}_{\mathcal{X}} $ and $\psi: \mathcal{Y} \mapsto \mathcal{H}_{\mathcal{Y}}$ such that $\kappa_{\mathcal{X}}(x,x') = \langle \phi(x),\phi(x') \rangle_{\mathcal{X}}$ and $\kappa_{\mathcal{Y}}(y,y') = \langle \psi(y),\psi(y') \rangle_{\mathcal{Y}}$. Assuming the normalization condition \eqref{eq:normalized_kernel_condition} for both $\kappa_{\mathcal{X}}$ and $\kappa_{\mathcal{Y}}$, we define the operator $Q : \mathcal{H}_{\mathcal{X}}\otimes\mathcal{H}_{\mathcal{Y}} \mapsto \mathcal{H}_{\mathcal{X}}\otimes\mathcal{H}_{\mathcal{Y}}$ as:
\begin{equation}\label{eq:operator_cross_bilinear_form}
\mathcal{Q}(f,g) = \langle f, Qg \rangle_{\otimes} = \int\limits_{\mathcal{X}}\int\limits_{\mathcal{Y}}\langle f,\phi_{\otimes}(x,y) \rangle_{\otimes} \langle \phi_{\otimes}(x,y),g \rangle_{\otimes} \mathrm{d}P_{\mathcal{X},\mathcal{Y}}(x,y).
\end{equation}
Notice that $f$ and $g$ belong to $\mathcal{H}_{\mathcal{X}}\otimes\mathcal{H}_{\mathcal{Y}}$, with reproducing kernel $\kappa_{\otimes}((x,y), (x',y')) = \kappa_{\mathcal{X}}(x,x')\kappa_{\mathcal{Y}}(y,y')$. The finite counterpart $\widehat{Q}_N$ can be defined in similar to \eqref{eq:empirical_version_G}.
\begin{prop}\label{thm:convergence_of_conditional} For positive definite kernels $\kappa_{\mathcal{X}}$ and $\kappa_{\mathcal{Y}}$ satisfying \eqref{eq:normalized_kernel_condition}, and $\kappa_{\mathcal{Y}}(y,y) = 1$ and $\kappa_{\mathcal{X}}(x,x) < C$. Then, with probability $1-\delta$
\begin{equation}\label{eq:bound_traces}
\left\vert \frac{\mathrm{tr}\left(G^\alpha \right)}{\mathrm{tr}\left(Q^\alpha \right)} - \frac{\mathrm{tr}\left(\widehat{G}_N^\alpha \right)}{\mathrm{tr}\left(\widehat{Q}_N^\alpha \right)} \right\vert \leq 2\beta \alpha C\sqrt{\frac{2\log{\frac{2}{\delta}}}{N}}, 
\end{equation}
where $\beta = 1/\mathrm{tr}\left(Q^\alpha \right)$.
\begin{prf}
Let  $a = \mathrm{tr}\left(Q^\alpha \right)$,  $\widehat{a} = \mathrm{tr}\left(\widehat{Q}_N^\alpha \right)$, $b = \mathrm{tr}\left(G^\alpha \right)$, and $\widehat{b} = \mathrm{tr}\left(\widehat{G}_N^\alpha \right)$.
\begin{eqnarray}
	\left\vert \frac{b}{a} - \frac{\widehat{b}}{\widehat{a}}	\right\vert & = & \left\vert \frac{\widehat{a}b - \widehat{b}a}{\widehat{a}a}	\right\vert \\
	& = & \left\vert \frac{b -\widehat{b}}{a} - \frac{\widehat{b}}{\widehat{a}}\frac{a^2 - \widehat{a}^2}{a(\widehat{a} + a)}	\right\vert. 
\end{eqnarray}
It follows from \eqref{eq:hadamard_conditional_entropy_inequality} that:
\begin{equation}\label{eq:bound_conditional_entropy_ratio}
\left\vert \frac{\mathrm{tr}\left(G^\alpha \right)}{\mathrm{tr}\left(Q^\alpha \right)} - \frac{\mathrm{tr}\left(\widehat{G}_N^\alpha \right)}{\mathrm{tr}\left(\widehat{Q}_N^\alpha \right)} \right\vert \leq  \frac{\vert b -\widehat{b} \vert}{a} + \frac{\vert a^2 - \widehat{a}^2 \vert}{a(\widehat{a} + a)}	
\end{equation}
Combining \eqref{eq:bound_traces} with \eqref{eq:bound_conditional_entropy_ratio} and noticing that $a^2 - \widehat{a}^2 = (a - \widehat{a})((a + \widehat{a}))$, yields the desired result.\\
$\square$
\end{prf}
\end{prop} 

% --------------------------------------------------------------------------
\section{Experiments}
\begin{figure}
\centering
\includegraphics[width = 9cm]{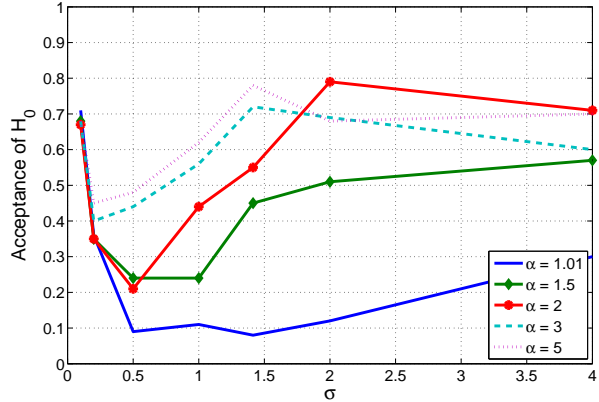}
\caption[Results of independence test under different parameters]{Results of the independence test based on the gap between tensor and  Hadamard product entropies for different kernel sizes $\sigma$ and entropy orders $\alpha$ for a fixed sample of size $1024$ and rotation angle $\theta = \frac{\pi}{8}$. The dimensionality of the of the random variables is $d=2$. }\label{fig:Indpendence_test_gap_MI_parameter} 
\end{figure}

%---------------------------------------------------------------------------
Experiments using the definition of conditional entropy for metric learning have been reported in \cite{LSanchez13}. 
Here, we develop a test for independence between random elements $X$ and $Y$ based on the gap between the entropy of the tensor and Hadamard products of their Gram matrices applied to an experimental setup similar to \cite{AGretton10}. We draw $N$ \emph{i.i.d.} samples from two randomly picked densities corresponding to the ICA benchmark densities \cite{FBach02}. These densities are scaled and shifted such that they have zero mean and unit variance (see Table \ref{tab:ICA_benchmark}).
\begin{table}
\caption{List of distributions used in the independence test along with their corresponding original and resulting kurtosis after centralization and rescaling}\label{tab:ICA_benchmark}
\centering
\begin{tabular}{l|c}
\textbf{Distribution} &\textbf{Kurtosis} \\ \hline \hline
Student's $t$ distribution $3$ DOF &  $\infty$ \\
Double exponential  &  $3.00$ \\
Uniform &  $-1.20$ \\ \hline
Student's $t$ distribution $5$ DOF &  $6.00$ \\
Exponential &  $6.00$  \\
Mixture, $2$ double exponentials &  $-1.16$ \\ \hline
Symmetric mixture, $2$ Gaussian, multimodal &  $-1.68$  \\
Symmetric mixture, $2$ Gaussian, transitional &  $-0.74$ \\
Symmetric mixture, $2$ Gaussian, unimodal &  $-0.50$  \\ \hline
Asymmetric mixture, $2$ Gaussian, multimodal &  $-0.53$ \\
Asymmetric mixture, $2$ Gaussian, transitional &  $-0.67$ \\
Asymmetric mixture, $2$ Gaussian, unimodal &  $-0.47$ \\ \hline
Symmetric mixture, $4$ Gaussian, multimodal &  $-0.82$  \\
Symmetric mixture, $4$ Gaussian, transitional &  $-0.62$ \\
Symmetric mixture, $4$ Gaussian, unimodal &  $-0.80$ \\ \hline
Asymmetric mixture, $4$ Gaussian, multimodal &  $-0.77$  \\
Asymmetric mixture, $4$ Gaussian, transitional &  $-0.29$  \\
Asymmetric mixture, $4$ Gaussian, unimodal &  $-0.67$ \\ 
\end{tabular}
\end{table}
The pair of random variables are mixed using a $2$-dimensional rotation matrix with rotation angle $\theta \in [0,\pi/4]$. Gaussian noise with unit variance and zero mean is a added as extra dimensions. Finally, each one of the random vectors is rotated by a random rotation (orthonormal matrix) in $\mathbb{R}^2$, and $\mathbb{R}^3$, accordingly. This causes the resulting random vectors to be dependent across all observed dimensions. We perform experiments varying angles, samples sizes, and dimensionality. The test compares the value of the gap:
\begin{equation}\label{eq:Indpendence_test_gap}
 S_{\alpha}(\mathbf{K}_X) + S_{\alpha}(\mathbf{K}_Y) - S_{\alpha}\left(\frac{\mathbf{K}_X \circ \mathbf{K}_Y}{\mathrm{tr}{(\mathbf{K}_X\circ \mathbf{K}_Y)}}\right),
\end{equation}
where $\mathbf{K}_X$ and $\mathbf{K}_Y$ are the Gram matrices (Gaussian kernel) for the $X$ and $Y$ components of the sample $\{(x_i,y_i)\}_{i=1}^N$, with a threshold computed by sampling a surrogate of the null hypothesis $H_0$ based on shuffling one of the components of the sample $k$ times, that is, the correspondences between $x_i$ and $y_i$ are broken by the random permutations. The threshold is the estimated quantile $1-\tau$ where $\tau$ is the significance level of the test (Type I error). The hypothesis $H_0$, $X$ is independent of $Y$, is accepted if the gap \eqref{eq:Indpendence_test_gap} is below the threshold, otherwise, we reject $H_0$. In all our experiments $k = 100$ and $\alpha = 1.01$.
\begin{figure*}
\centering
\subfigure[]{\includegraphics[height=4cm,width = 5.9cm]{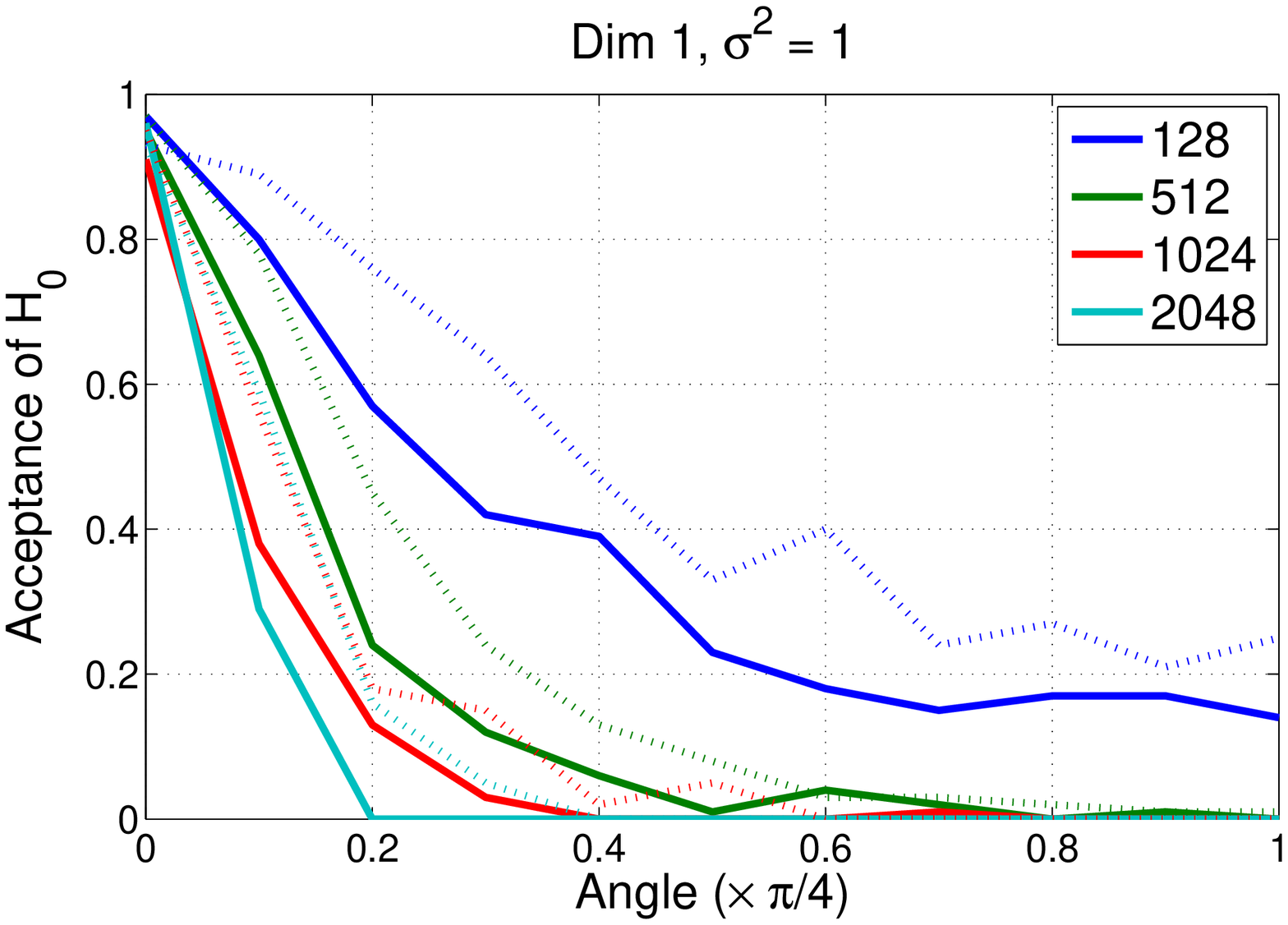}\label{fig:Independence_test_MI_d1}}
\subfigure[]{\includegraphics[height=4cm,width = 5.9cm]{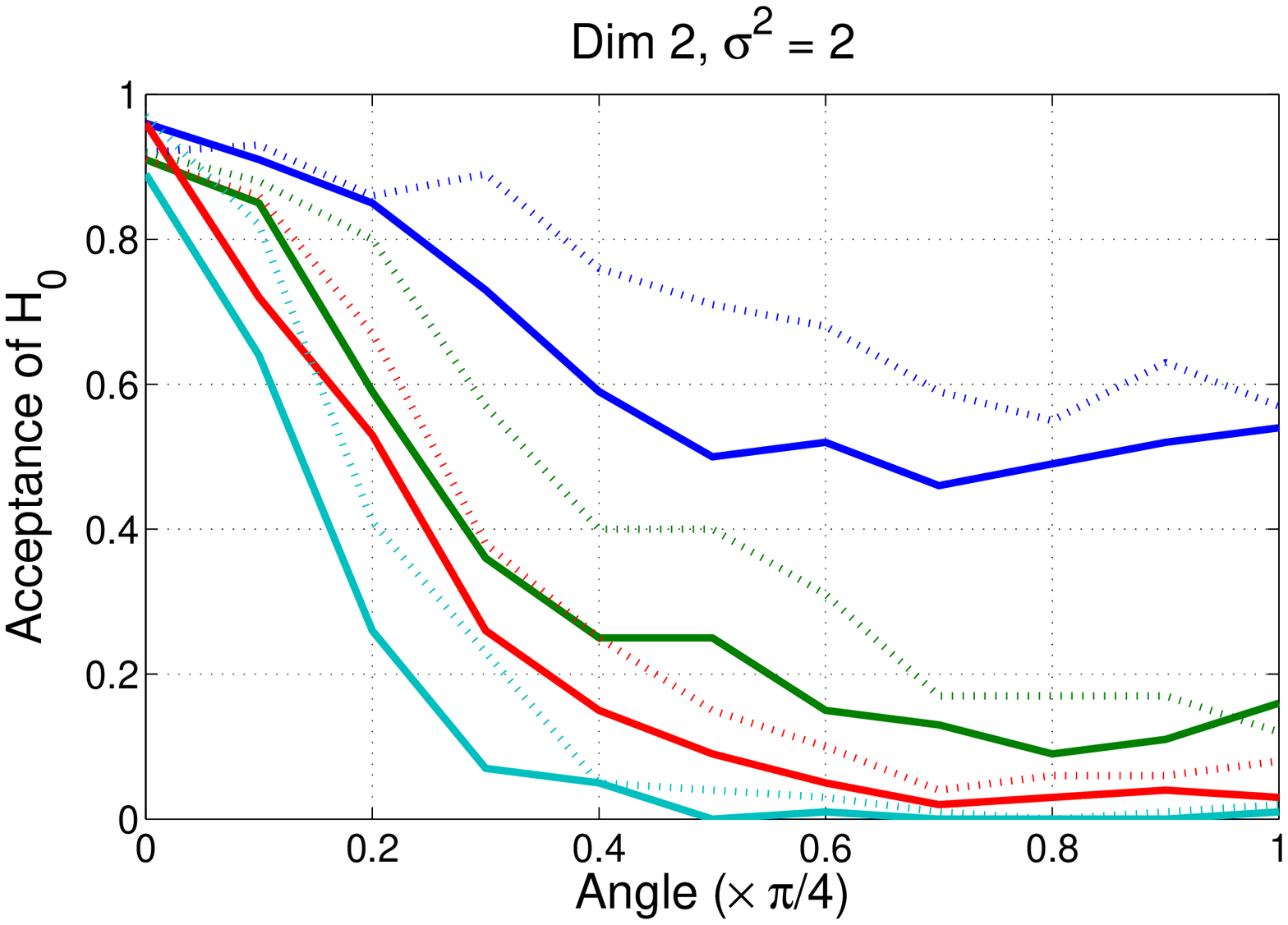}\label{fig:Independence_test_MI_d2}}
\subfigure[]{\includegraphics[height=4cm,width = 5.9cm]{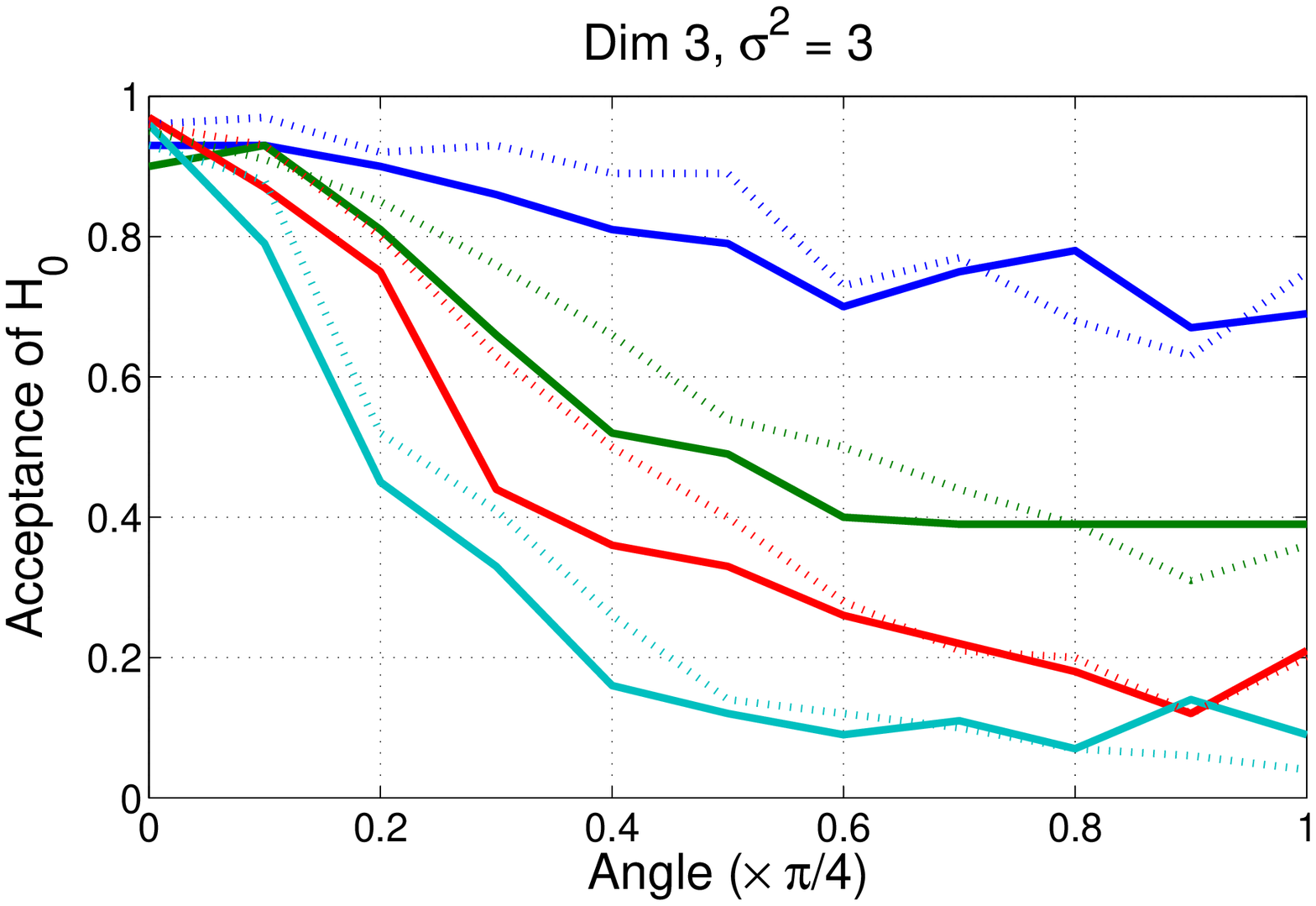}\label{fig:Independence_test_MI_d3}}\\
\subfigure[]{\includegraphics[height=4cm,width = 5.9cm]{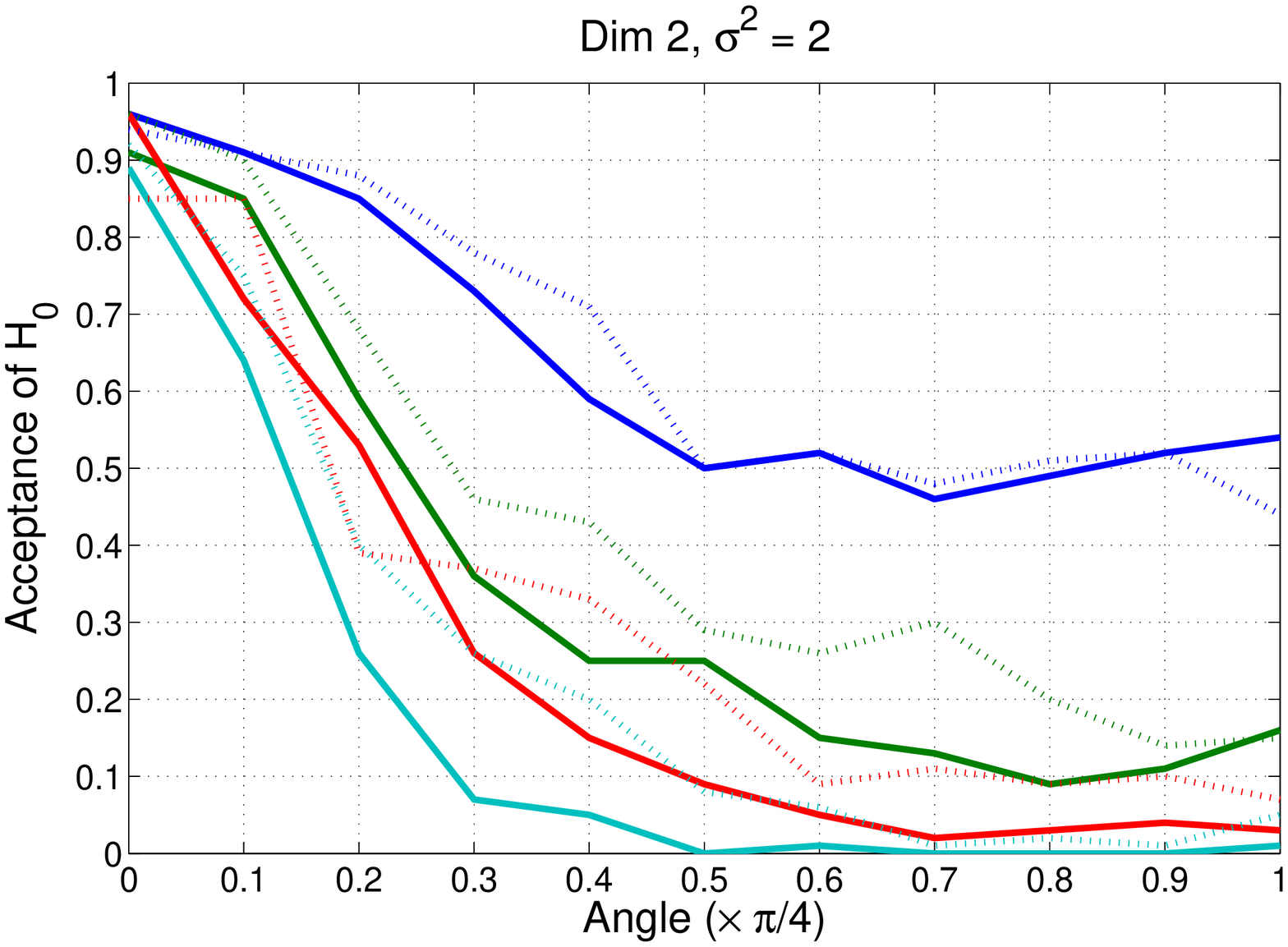}\label{fig:Independence_test_MST_d2}}
\subfigure[]{\includegraphics[height=4cm,width = 5.9cm]{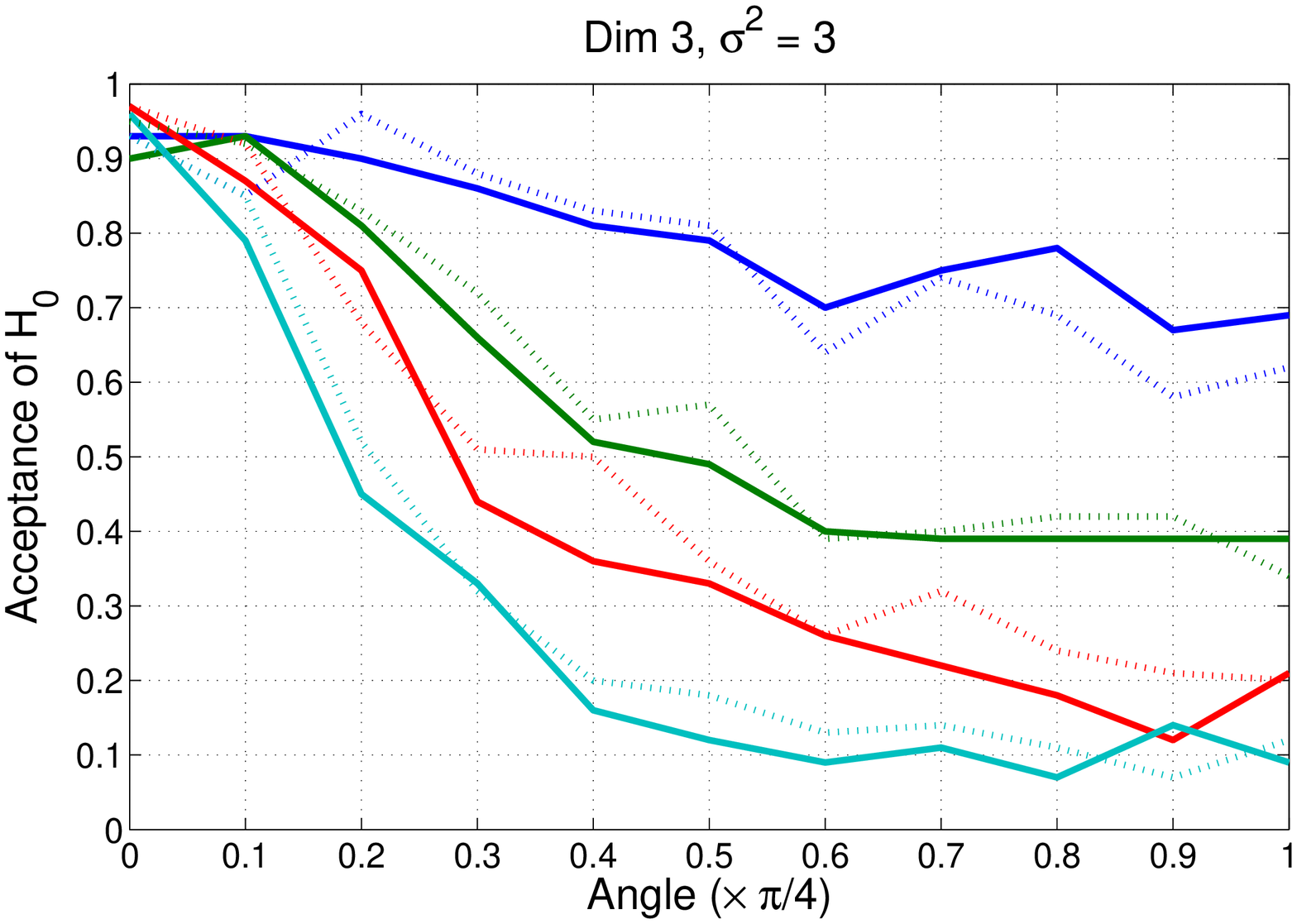}\label{fig:Independence_test_MST_d3}}
\caption{Results of the independence test based on the gap between tensor and  Hadamard product entropies for different sample sizes and dimensionality. Figures \ref{fig:Independence_test_MI_d1}, \ref{fig:Independence_test_MI_d2}, and\ref{fig:Independence_test_MI_d3}, correspond to the estimated acceptance rates for $H_0$  for random  variables of $1$, $2$, and $3$ dimensions, and compare the results between the proposed test (solid lines) and the kernel-based statistic (dotted) proposed in \cite{AGretton10}. Figures  \ref{fig:Independence_test_MST_d2}, and\ref{fig:Independence_test_MST_d3}, correspond to the estimated acceptance rates for $H_0$  for random  variables of $2$, and $3$ dimensions, and compare the results between the proposed test (solid lines) and the minimum spanning graph entropy estimator (dotted) proposed in \cite{AHero98}.The larger the angle the easier to reject independence $H_0$.}\label{fig:Indpendence_test_gap_MI} 
\end{figure*}
The solid lines in Figures \ref{fig:Independence_test_MI_d1}, \ref{fig:Independence_test_MI_d2}, \ref{fig:Independence_test_MI_d3}, \ref{fig:Independence_test_MST_d2}, and \ref{fig:Independence_test_MST_d3} show the estimated probability of $H_0$ being accepted for the proposed test with $\tau = 0.05$. The dotted lines in Figures \ref{fig:Independence_test_MI_d1}, \ref{fig:Independence_test_MI_d2}, \ref{fig:Independence_test_MI_d3}  are the acceptance rates obtained using the kernel-based statistic proposed in \cite{AGretton10},
\begin{equation}\label{eq:gretton_statistic}
\begin{split}
T_n = & \frac{1}{n^2}\sum_{i,j=1}^n L_h(x_i-x_j)L_h^{\prime}(y_i,y_j)+\\
 & -\frac{2}{n^3}\sum_{j = 1}^n\left[\left(\sum_{i=1}^nL_h(x_i-x_j)\right)\left(\sum_{i=1}^nL_h^{\prime}(y_i-y_j)\right)\right] +\\
 & +\left(\frac{1}{n^2}\sum_{i,j=1}^nL_h(x_i-x_j)\right)\left(\frac{1}{n^2}\sum_{i,j=1}^nL_h^{\prime}(y_i-y_j)\right),
\end{split}
\end{equation}
where $L_h$ and $L_h^{\prime}$ are characteristic kernels on $\mathbb{R}^d$ \cite{BSriperumbudur08}. The dotted lines in Figures  \ref{fig:Independence_test_MST_d2}, \ref{fig:Independence_test_MST_d3} are the acceptance rates for an statistic based on the difference between joint and marginal Renyi's entropies estimated using the minimum spanning tree graph as proposed in \cite{AHero98}, namely,
\begin{equation}\label{eq:entropic_graph_statistic}
\tilde{H}_{\alpha}(X_n)+\tilde{H}_{\alpha}(Y_n) - \tilde{H}_{\alpha}(X_n,Y_n), 
\end{equation}
where $\tilde{H}_{\alpha}(Z_n) = \frac{1}{1-\alpha}\log{\min\limits_{e \in \mathcal{T}} \sum\limits_{e}\vert e\vert^{\gamma}}$. $\mathcal{T}$ is the set of vertices of the entropic graph and $\vert e\vert$ denotes the Euclidean norm of the edge $e$. Notice that here we don't consider the bias-correction constant that is presented in \cite{AHero98} since it will be present on the threshold as well. The results are averages over $100$ simulations for each one of the parameter configurations. In the case of $X,Y \in \mathbb{R}$ (Figure \ref{fig:Independence_test_MI_d1}), the type II error is low even for small sample sizes, whereas the dependence becomes more difficult to detect as $d$ increases, requiring a larger $N$ to obtain an acceptable type II error. Our results are competitive to those obtained with the kernel based statistic \eqref{eq:gretton_statistic} and the entropic graph estimator \eqref{eq:entropic_graph_statistic}. The three methods perform relatively similar for large angles, but it can be noticed that the proposed method works better when the angle is close to $0$.  It is important to point out that in all cases (the proposed statistic using the gap, the one in \eqref{eq:gretton_statistic}, and \eqref{eq:entropic_graph_statistic}) the threshold was empirically determined by approximating the null distribution using permutations on one of the variables. Whether we can provide a distribution of the null hypothesis for \eqref{eq:Indpendence_test_gap} is subject of future work. Figure \ref{fig:Indpendence_test_gap_MI_parameter} shows the influence of the parameters in the power of the proposed independence test. 
The behavior of the test for different orders $\alpha$ and kernel sizes $\sigma$ can be explained from the spectral properties of the Gram matrices. For smaller kernel sizes the Gram matrix approaches to identity and thus its eigenvalues become more similar, with $1/n$ as the limit case. Therefore, the gap \eqref{eq:Indpendence_test_gap}  monotonically increases as $\sigma \rightarrow 0$, so does the gap for the permuted sample. Since both quantities have the same upper bound, the probability of accepting $H_0$ increases. The other phenomenon is related to the entropy order, it can be noticed that the larger the order $\alpha$ the smaller the kernel size $\sigma$ that is needed to minimize the type II error. The order has an smoothing effect in the resulting operator defined in \eqref{eq:any_order_operator_definition}. Large $\alpha$ will emphasize on the largest eigenvalues of the Gram matrices that are commonly associated with slowly changing features.  

%-------------------------------------------------------------------------------------------------------- 
\section{Conclusions}
%-------------------------------------------------------------------------------------------------------- 
We presented an estimation framework of entropy-like quantities based on infinitely divisible matrices. By using the axiomatic characterization of Renyi's entropy, a functional on positive definite matrices is defined. This functional resembles the definitions of entropy in quantum information theory; however, our analysis is different from quantum information theory since we need to consider not only the functional but also the kernel employed to compute the proposed measures of entropy. The use of Hadamard products allows us to define quantities that are similar to mutual information and conditional entropy, and set the conditions that lead to infinitely divisibility. We showed some properties of the proposed quantities and their asymptotic behavior as operators in reproducing kernel Hilbert spaces defined by distribution-dependent kernels. An important result is that the convergence of the eigenvalues of the Gram matrix to the eigenvalues of the integral operators is independent of the input dimensionality. Numerical experiments showed the usefulness of the proposed approach for independence testing with results that compete with the state of the art. 

%% Appendixes
\begin{appendices}
\section{Auxiliary Results}
\subsection{Hilbert Space representation of Data}
We want to show that the function $d(x,y)$ defined in \eqref{eq:induced_metric_hilbert_representation} is a semi-metric. 
Since $G(\cdot, \cdot)$ is a positive definite function,
\begin{equation}
0 \leq \det{\left(\begin{array}{cc}
G(x,x) & G(x,y) \\
G(y,x) & G(y,y)  
\end{array}\right)} = G(x,x)G(y,y) - G(x,y)G(y,x),
\end{equation} 
which yields,
\begin{equation}
0 \leq G(x,x) + G(y,y) - 2\sqrt{G(x,x)G(y,y)} \leq G(x,x) + G(y,y) - 2G(x,y) = d^2(x,y).
\end{equation}
Symmetry of $d(x,y)$ follows immediately from the symmetry of $G(x,y)$.
By the Cauchy-Schwarz inequality,
\begin{eqnarray}\label{eq:CS-inequality}
G(x,z)+G(y,z)-G(x,y)-G(z,z) & \leq & \left\vert \int\limits_{\mathcal{T}}(\phi(t,x)-\phi(t,z))(\phi(t,z)-\phi(t,y))\mathrm{d}\mu_{\mathcal{T}}(t) \right\vert \\ 
\nonumber & \leq &  \sqrt{\int\limits_{\mathcal{T}}(\phi(t,x)-\phi(t,z))^2\mathrm{d}\mu_{\mathcal{T}}(t)\int\limits_{\mathcal{T}}(\phi(t,y)-\phi(t,z))^2\mathrm{d}\mu_{\mathcal{T}}(t)}\\
\nonumber & = & \sqrt{d^2(x,z)d^2(y,z)} \\
& = & d(x,z)d(y,z)
\end{eqnarray}

\begin{equation}\label{eq:app_triangle_inequality}
\begin{split}
d^2(x,y) = &\:  G(x,x)+G(y,y)-2G(x,y)\\
  = &\: G(x,x)+G(z,z)-2G(x,z)+G(y,y)+G(z,z)-2G(y,z)+ \\
 &\:\:\: + 2\left[G(x,z)+G(y,z)-G(x,y)-G(z,z)\right] \\ 
 \leq &\: d^2(x,z) + d^2(y,z) + 2d(x,z)d(y,z)\\
  = &\: \left(d(x,z) + d(y,z)\right)^2
\end{split}
\end{equation}
\subsection{Functional Calculus on Hermitian Matrices}\label{app:spectral_theorem}
The following spectral decomposition theorem relates to the functional calculus on matrices and provides a reasonable way to extend continuous scalar-valued functions to Hermitian matrices.  
\begin{thm}\label{thm:functional_calculus_psd_matrix} Let $D\subset \mathbb{C}$ be a given set and let $\mathcal{N}_n(D) := \{A \in M_n:\:A \textrm{ is normal and }\: \sigma(A)\subset D\}$. If $f(t)$ is a continuous scalar-valued function on $D$, then the primary matrix function
\begin{equation}\label{eq:primary_matrix_function}
f(A) = U \left(\begin{array}{ccc}
f(\lambda_1) & \cdots & 0 \\
\vdots & \ddots & \vdots \\
0 & \cdots & f(\lambda_n) \\
\end{array}\right)U^*
\end{equation}
is continuous on $\mathcal{N}_n(D)$, where $A = U \Lambda U^*$, $\Lambda = diag(\lambda_1,\dots,\lambda_n)$, and $U \in M_n$ is unitary.
\end{thm}
\section{Non-Negative Sequences and Renyi's Entropy Axioms}\label{app:Renyi_Axioms}
\begin{figure}
\centering
\includegraphics[width = 8cm]{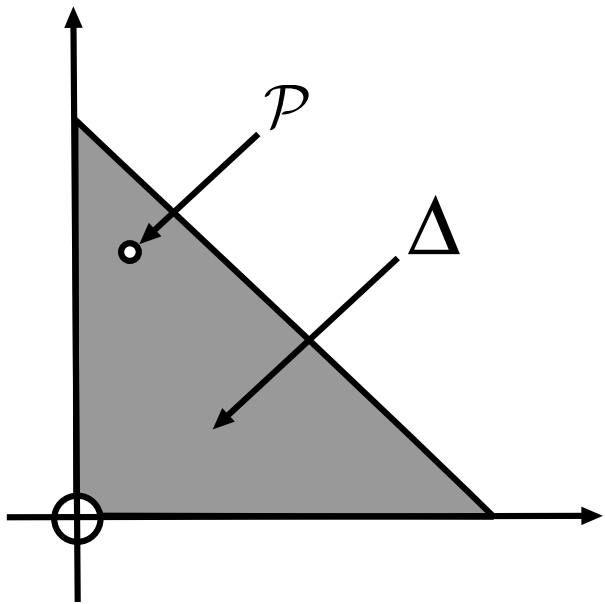} 
\caption{The generalized probability space}\label{fig:gen_probability_space}
\end{figure}
For simplicity of the exposition, we will only consider the discrete case were the sample space $\Omega$ is countable (not necessarily finite). Consider a sequence $\mathcal{P} = \{p_i\}_{i=1}^n$ of non-negative numbers and a function $W(\mathcal{P}) = \sum p_i$  such that $0 < W(\mathcal{P}) \leq 1$. The sequence $\mathcal{P}$ is called a generalized probability distribution and $\Delta_n$ the space of all generalized probability distributions up to dimension $n$. The entropy $H(\mathcal{P})$ of a generalized probability distribution (see Figure \ref{fig:gen_probability_space}) is characterized by the following postulates:
\begin{enumerate}[I.]
\item \label{property:symmetry} $H(\mathcal{P})$ is a symmetric function of the elements of $\mathcal{P}$
\item \label{property:continuity} If $\{p\}$ is single element generalized probability distribution then $H(p)$ is a continuous function for $0 < p \leq 1$.
\item \label{property:max_localization} $H(\frac{1}{2}) = 1$.
\item \label{property:joint} If $\mathcal{P} \in \Delta_n$ and $\mathcal{Q} \in \Delta_n$, $H(\mathcal{P} \otimes \mathcal{Q}) = H(\mathcal{P}) + H(\mathcal{Q})$.
\item \label{property:generalized_mean_value} There exists a strictly monotonic and continuous function $g$ such that if $\mathcal{P} \in \Delta_n$ and $\mathcal{Q} \in \Delta_n$ and $W(\mathcal{P})+ W(\mathcal{Q}) \leq 1$, we have:
\begin{equation}\label{eq:entropy_meanvalue}
H(\mathcal{P}\odot \mathcal{Q}) = g^{-1}\left(\frac{W(\mathcal{P})g(H(\mathcal{P}))+W(\mathcal{Q})g(H(\mathcal{Q}))}{W(\mathcal{P})+W(\mathcal{Q})}\right)
\end{equation} 
Here, $\mathcal{P}\odot \mathcal{Q}$ denotes the concatenation or union of the sequences $\mathcal{P}$ and $\mathcal{Q}$. 
\end{enumerate}
Notice that \eqref{eq:entropy_meanvalue} is a generalization of the mean value property which is usually associated with the interpretation of entropy as the average uncertainty. The function $H_\alpha(\mathcal{P})$ defined as
\begin{equation}\label{eq:renyi_alpha_entropy}
H_{\alpha}(\mathcal{P}) = \frac{1}{1-\alpha}\log_{2}{\left(\frac{\sum_{i}p_i^{\alpha}}{\sum_{i}p_i}\right)},
\end{equation}
where $g(x) = g_\alpha(x) = 2^{(\alpha-1)x}$, satisfies conditions \ref{property:symmetry}-\ref{property:generalized_mean_value} for $\alpha >0$ and $\alpha \neq 1$. The above function \eqref{eq:renyi_alpha_entropy} is known as the Rényi's entropy of order $\alpha$. 
An important feature of this functional is that is monotonically increasing on a partial ordering defined on the $\sigma$-algebras called refinement. In short, let $(\Omega,\mathcal{B})$ be a measurable space and let $\mathcal{B}_1$ and $\mathcal{B}_2$ be two sub-$\sigma$-algebras of $\mathcal{B}$, we say $\mathcal{B}_1 \prec \mathcal{B}_2$ if $\mathcal{B}_1 \subset \mathcal{B}_2$. 

\begin{figure}
\centering
\includegraphics[width = 8cm]{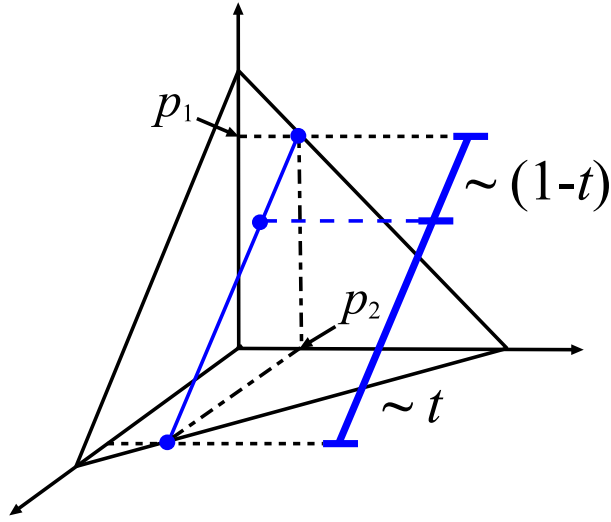} 
\caption{Refinements and the generalized probability space}\label{fig:simplex_generalized_distributions}
\end{figure}

In the case of Shannon's entropy, conditions \ref{property:joint} and \ref{property:generalized_mean_value} are usually summarized by the following:
\begin{itemize}
\item For a any distribution $\mathcal{P} = (p_1,p_2,\dots,p_n)$, $H(tp_1,(1-t)p_1,p_2,\dots,p_n) = H(\mathcal{P})+p_1H(t,1-t)$ for $0 \leq t \leq 1$.
\end{itemize}
Moreover, notice that the concatenation or union of sequences can be simply represented by a constrained linear combination argument. Consider two generalized distributions $\mathcal{P}$ and $\mathcal{Q}$ in $\Delta_n$ such that $W(\mathcal{P})+W(\mathcal{Q})\leq 1$, the operation $\mathcal{P}\odot \mathcal{Q}$ is equivalent to $\mathcal{P}+\mathcal{Q} = \{p_i+q_i\}_{i=1}^n$ if $\sum_{i=1}^n p_iq_i = 0$. Figure \ref{fig:simplex_generalized_distributions} depicts the above statement in a simple example, that combine the idea of a refinement and the orthogonality condition. Consider a distribution $\mathcal{P} = (p_1,p_2)$ and a refinement $\mathcal{P}_r = (tp_1,(1-t)p_1,p_2)$. One possible way to look at this refinement is as the result of a concatenation of two generalized distributions $(tp_1,(1-t)p_1)$  and $(p_2)$ as depicted by the plane normal to the direction of $p_2$ cutting the simplex. An analogue interpretation for the operators is the condition $AB = BA = \mathbf{0}$ in Proposition \ref{prop:matrix_entropy}. 
\section{Infinitely Divisible Kernels}
%--------------------------------------------------------------------------------------------------------------------------------------
\subsection{Direct-Sum and Product kernels}
\subsubsection{Direct-Sum kernels}
Let $\kappa_1$ and $\kappa_2$ be two positive definite kernels defined on $\mathcal{X} \times \mathcal{X}$. The kernel $\kappa_{\oplus} = \kappa_1+ \kappa_2$, defined as $\kappa_{\oplus}(x,y) = \kappa_1(x,y) + \kappa_2(x,y)$, is a positive definite kernel. The above function is called direct sum kernel and it is the reproducing kernel of a space $\mathcal{H}_{\oplus}$ of functions of the form $f = f_1 + f_2$, where $f_1 \in \mathcal{H}_1$ and $f_2 \in \mathcal{H}_2$, and $\mathcal{H}_1$ and $\mathcal{H}_2$ are the RKHSs defined by $\kappa_1$ and $\kappa_2$, respectively. Consider the Hilbert space $\mathcal{H} = \mathcal{H}_1\times \mathcal{H}_2$ formed by all pairs $(f_1,f_2)$ coming from $\mathcal{H}_1$ and $\mathcal{H}_2$,respectively. It is possible that some functions $f \neq 0 $ belong to both $\mathcal{H}_1$ and $\mathcal{H}_2$ at the same time. These functions form a set of pairs $(f,-f) \in \mathcal{H}$, which turn out to be a closed subspace of $\mathcal{H}$ denoted by $\mathcal{H}_0$, such that, $\mathcal{H} = \mathcal{H}_0 \oplus \mathcal{H}_0^{\perp}$. Therefore, the linear correspondence $f(x) = f_1(x) + f_2(x)$ between $f \in \mathcal{H}_{\oplus}$ and $(f_1,f_2) \in \mathcal{H}$ is such that all elements in $\mathcal{H}_0$ map to the zero function in $\mathcal{H}_{\oplus}$ and the elements of $\mathcal{H}_{\oplus}$ and $\mathcal{H}_0^{\perp}$ are in one to one correspondence. The norm of $f \in \mathcal{H}_{\oplus}$ can be defined from the correspondence $f \mapsto (g_1(f),g_2(f))$ as:
\begin{equation}\label{eq:direct_sum_correspondence}
\Vert f \Vert_{\mathcal{H}\oplus}^{2} = \Vert(g_1(f),g_2(f))\Vert_{\mathcal{H}}^{2} = \Vert g_1(f)\Vert_{\mathcal{H}_1}^{2} +\Vert g_2(f)\Vert_{\mathcal{H}_2}^{2}
\end{equation}
Notice that, $(g_1(f),g_2(f))$ is the decomposition of $f$ into the pair $\mathcal{H}$ with minimum norm in this space. The following theorem states the result \cite{NAronszajn50}.
\begin{thm}\label{thm:direct_sum_kernel}
If $\kappa_i(x,y)$ is the reproducing kernel of the class $\mathcal{H}_i$, with norm $\Vert \cdot \Vert_i$, then $\kappa(x,y) = \kappa_1(x,y) + \kappa_2(x,y)$ is the reproducing kernel of the class of functions $\mathcal{H}_{\oplus}$ of all functions $f = f_1+f_2$ with $f_i \in \mathcal{H}_i$, and with the norm defined by
\begin{equation}
\Vert f \Vert_{\oplus}^2 = \min \left\{ \Vert f_1 \Vert_{1}^2 + \Vert f \Vert_{2}^2 \right\}.
\end{equation}
The minimum is taken over all decompositions $f = f_1+f_2$ with $f_i \in \mathcal{H}_i$
\end{thm} 
\subsubsection{Product Kernel and Tensor Product Spaces}
Consider two positive definite kernels $\kappa_1$ and $\kappa_2$ defined on $\mathcal{X} \times \mathcal{X}$ and $\mathcal{Y} \times \mathcal{Y}$, respectively. Their tensor product $\kappa_1 \otimes \kappa_2: (\mathcal{X} \times \mathcal{Y}) \times (\mathcal{X} \times \mathcal{Y})$ defined by:
\begin{equation}\label{eq:kernel_tensor_product} 
 \kappa_1 \otimes \kappa_2((x_i,y_i),(x_j,y_j)) =  \kappa_1(x_i,x_j)\kappa_2(y_i,y_j)
\end{equation}
is also a positive definite kernel. Note that we can consider two kernels $\tilde{\kappa}_1$ and $\tilde{\kappa}_2$, both defined on $(\mathcal{X} \times \mathcal{Y}) \times (\mathcal{X} \times \mathcal{Y})$, such that $ \tilde{\kappa}_1((x_i,y_i),(x_j,y_j)) = \kappa_1(x_i,x_j)$ and $\tilde{\kappa}_2((x_i,y_i),(x_j,y_j)) = \kappa_1(y_i,y_j)$; the kernel $\tilde{k}_1\cdot\tilde{\kappa}_2((x_i,y_i),(x_j,y_j))$ 
\[
\begin{split} 
& = \tilde{\kappa}_1((x_i,y_i),(x_j,y_j))\tilde{\kappa}_2((x_i,y_i),(x_j,y_j)),\\ 
&=  \kappa_1 \otimes \kappa_2((x_i,y_i),(x_j,y_j)),
\end{split}\]
and is positive definite by the Schur Theorem. 
Let us look at the space of functions that $\kappa_{\otimes} = \kappa_1 \otimes \kappa_2$ spans. Let $\mathcal{H}_{\otimes} = \mathcal{H}_1 \otimes \mathcal{H}_2$, where $\mathcal{H}_1$ and $\mathcal{H}_2$ are the RKHSs spanned by $\kappa_1$ and $\kappa_2$, respectively. The space $\mathcal{H}_{\otimes}$ is the completion of the space of all functions $f$ on $\mathcal{X}\times \mathcal{Y}$ of the form:
\begin{equation}\label{eq:tensor_representation}
f(x,y) = \sum\limits_{i=1}^{n}f_1^{(i)}(x)f_2^{(i)}(y)
\end{equation} 
with $f_1^{(i)} \in \mathcal{H}_1$ and $f_2^{(i)} \in \mathcal{H}_2$, and inner product,
\begin{equation}\label{eq:tensor_inner_product}
\langle f,g \rangle_{\otimes} = \sum\limits_{i=1}^{n}\sum\limits_{j=1}^{m}\langle f_1^{(i)},g_1^{(j)}\rangle_1 \langle f_2^{(i)},g_2^{(j)}\rangle_2.
\end{equation}
The functions $f$ and $g$ may have multiple representations of the form \eqref{eq:tensor_representation} without changing $\langle f,g \rangle_{\otimes}$. Let us look at the case where $\mathcal{X}$ and $\mathcal{Y}$ are the same set. The following theorem describes the kernel derived from the restriction of $\kappa_1 \otimes \kappa_2$ to the diagonal subset of $\mathcal{X}\times\mathcal{X}$ \cite{NAronszajn50}.
\begin{thm}\label{eq:direct_product_kernel_diagonal}
For $x,y \in \mathcal{X}$, the kernel$\kappa(x,y) = \kappa_1(x,y)\kappa_2(x,y)$ is the reproducing kernel of the class $\mathcal{H}$ of the restrictions of the direct product $\mathcal{H}_{\otimes} = \mathcal{H}_1 \otimes \mathcal{H}_2$ to the diagonal set formed by all elements $(x,x) \in \mathcal{X}\times \mathcal{X}$. For any such restriction $f$, $\Vert f \Vert = \min \Vert g \Vert_{\otimes}$ for all $g \in\mathcal{H}_{\otimes}$ such that $f(x) = g(x,x)$.
\end{thm}
\subsection{Negative Definite Functions and Infinitely Divisible Matrices}
\subsubsection{Negative Definite Functions and Hilbertian Metrics}
Let $\mathcal{M} = \left( \mathcal{X},d \right)$ be a separable metric space, a necessary and sufficient condition for $\mathcal{M}$ to be embeddable in a Hilbert space $\mathcal{H}$ is that for any set $\{ x_i \} \subset \mathcal{X}$ of $n+1$ points, the following inequality holds:
\begin{equation}\label{eq:Hilbert_embeddable}
\sum\limits_{i,j =1}^n\alpha_i\alpha_j\left(d^2(x_0,x_i) + d^2(x_0,x_j) - d^2(x_i,x_j) \right) \geq 0,
\end{equation} 
for any $\bm{\alpha} \in \mathbb{R}^n$. This condition is equivalent to 
\begin{equation}\label{eq:condtionally_negative_definite}
\sum\limits_{i,j = 0}^n\alpha_i\alpha_j d^2(x_i,x_j)  \leq 0,
\end{equation}
for any $\bm{\alpha} \in \mathbb{R}^{n+1}$, such that $\sum_{i=0}^n\alpha_i = 0$. This condition is known as negative definiteness. Interestingly, the above condition implies that $\exp(-r d^2(x_i,x_j))$ is positive definite in $\mathcal{X}$ for all $r>0$ \cite{ISchoenberg38}. Indeed, matrices derived from functions satisfying the above property conform a special class of matrices know as infinitely divisible.
\end{appendices}
%--------------------------------------------------------------------------------------------------------------------------------------
\bibliographystyle{IEEEtran}
\bibliography{biblio}

% Generated by IEEEtran.bst, version: 1.13 (2008/09/30)
\begin{thebibliography}{10}
\providecommand{\url}[1]{#1}
\csname url@samestyle\endcsname
\providecommand{\newblock}{\relax}
\providecommand{\bibinfo}[2]{#2}
\providecommand{\BIBentrySTDinterwordspacing}{\spaceskip=0pt\relax}
\providecommand{\BIBentryALTinterwordstretchfactor}{4}
\providecommand{\BIBentryALTinterwordspacing}{\spaceskip=\fontdimen2\font plus
\BIBentryALTinterwordstretchfactor\fontdimen3\font minus
  \fontdimen4\font\relax}
\providecommand{\BIBforeignlanguage}[2]{{%
\expandafter\ifx\csname l@#1\endcsname\relax
\typeout{** WARNING: IEEEtran.bst: No hyphenation pattern has been}%
\typeout{** loaded for the language `#1'. Using the pattern for}%
\typeout{** the default language instead.}%
\else
\language=\csname l@#1\endcsname
\fi
#2}}
\providecommand{\BIBdecl}{\relax}
\BIBdecl

\bibitem{JPrincipe}
J.~C. Principe, \emph{Information Theoretic Learning: Renyi's Entropy and
  Kernel Perspectives}, ser. Series in Information Science and Statistics,
  M.~Jordan, R.~Nowak, and B.~Schölkopf, Eds.\hskip 1em plus 0.5em minus
  0.4em\relax Springer, 2010.

\bibitem{ARenyi60}
A.~Rényi, ``On measures of entropy and information,'' in \emph{Proceeding of
  the Fourth Berkeley Symposium on Mathematical Statistics and Probability},
  J.~Neyman, Ed., vol.~1.\hskip 1em plus 0.5em minus 0.4em\relax University of
  California Press, June 1960, pp. 547--561.

\bibitem{JWXu08}
J.-W. Xu, A.~R.~C. Paiva, I.~Park, and J.~C. Principe, ``A reproducing kernel
  hilbert space framework for information theoretic learning,'' \emph{IEEE
  Transactions on Signal Processing}, vol.~56, no.~12, pp. 5891--5902, December
  2008.

\bibitem{LSanchezGiraldo11}
L.~G. Sanchez-Giraldo and J.~C. Principe, ``A reproducing kernel hilbert space
  formulation of the principle of relevant information,'' in \emph{IEEE
  International Workshop on Machine Learning for Signal Processing (MLSP)},
  2011.

\bibitem{AHero98}
A.~Hero and O.~Michel, ``Robust entropy estimation strategies based on edge
  weighted random graphs,'' in \emph{Proceedings of the Meeting of the
  International Society for Optical Engineering (SPIE)}, July 1998.

\bibitem{AHero06}
J.~A. Costa and A.~O.~H. III, \emph{Statistics and Analysis of Shapes}, ser.
  Modeling and Simulation in Science, Engineering and Technology.\hskip 1em
  plus 0.5em minus 0.4em\relax Springer, 2006, ch. Determining Intrinsic
  Dimension and Entropy of High-Dimensional Shape Spaces, pp. 231--252.

\bibitem{NLeonenko08}
N.~Leonenko, L.~Pronzato, and V.~Savani, ``A class of rényi information
  estimators for multidimensional densities,'' \emph{The Annals of Statistics},
  vol.~36, no.~5, pp. 2153--2182, 2008.

\bibitem{BPoczos10}
D.~Pál, B.~Póczos, and C.~Szepesvári, ``Estimation of rényi entropy and mutual
  information based on generalized nearest-neighbor graphs.'' in \emph{NIPS},
  2010.

\bibitem{LFaivishevsky08}
L.~Faivishevsky and J.~Goldberger, ``Ica based on a smooth estimation of the
  differential entropy,'' in \emph{NIPS}, 2008, pp. 433--440.

\bibitem{MAizerman64_perceptron}
M.~A. Aizerman, E.~M. Braverman, and L.~I. Rozonoer, ``Theoretical foundations
  of the potential function method in pattern recognition learning,''
  \emph{Avtomatika i Telemekhanika}, vol.~25, no.~6, pp. 917--936, June 1964.

\bibitem{MAizerman64_klms}
------, ``The method of the potential functions for the problem of restoring
  the characteristic of a function converter from randomly observed points,''
  \emph{Avtomatika i Telemekhanika}, vol.~25, no.~12, pp. 1705--1714, December
  1964.

\bibitem{JShaweTaylor}
J.~Shawe-Taylor and N.~Cristianini, \emph{Kernel Methods for Pattern
  Analysis}.\hskip 1em plus 0.5em minus 0.4em\relax Cambridge University Press,
  2004.

\bibitem{FBach02}
F.~R. Bach and M.~I. Jordan, ``Kernel independent component analysis,''
  \emph{Journal of Machine Learning Research}, vol.~3, pp. 1--48, July 2002.

\bibitem{AGretton05}
A.~Gretton, O.~Bousquet, A.~Smola, and B.~Schölkopf, ``Measuring statistical
  dependence with hilbert-schmidt norms,'' in \emph{Proceedings of Algorithmic
  Learning Theory}, S.~Jain, H.~Simon, and E.~Tomita, Eds., 2005, pp. 63--77.

\bibitem{BSriperumbudur08}
B.~K. Sriperumbudur, A.~Gretton, K.~Fukumizu, G.~Lanckriet, and B.~Schölkopf,
  ``Injective hilbert space embeddings of probability measures,'' in
  \emph{Proceedings of the 21st Annual Conference on Learning Theory}, 2008,
  pp. 111--122.

\bibitem{SSeth11}
S.~Seth, M.~Rao, I.~Park, and J.~C. Príncipe, ``A unified framework for
  quadratic measures of independence,'' \emph{IEEE Transactions on Signal
  Processing}, vol.~59, no.~8, pp. 3624--3635, August 2011.

\bibitem{EParzen59}
E.~Parzen, ``Statistical inference on time series by hilbert space methods,
  i,'' Stanford University, Tech. Rep.~23, January 1959.

\bibitem{NAronszajn50}
N.~Aronszajn, ``Theory of reproducing kernels,'' \emph{Transactions of the
  American Mathematical Society}, vol.~68, no.~3, pp. 337--404, May 1950.

\bibitem{RKondor04}
T.~Jebara, R.~Kondor, and A.~Howard, ``Probability product kernelsx,''
  \emph{Journal of Machine Learning Research}, vol.~5, pp. 819--844, July 2004.

\bibitem{OBousquet04}
M.~Hein and O.~Bousquet, ``Hilbertian metrics and positive definite kernels on
  probability measures,'' Max Planck Institute for Biological Cybernetics,
  Tech. Rep. 126, July 2004.

\bibitem{AMartins09}
A.~F. Martins, N.~A. Smith, E.~P. Xing, P.~M.~Q. Aguilar, and M.~A.~T.
  Figueiredo, ``Nonextensive information theoretic kernels on measures,''
  \emph{Journal of Machine Learning Research}, vol.~10, pp. 935--975, April
  2009.

\bibitem{MOhya}
M.~Ohya and D.~Petz, \emph{Quantum Entropy and Its Use}, ser. Text and
  Monographs in Physics, R.~Balian, W.~Beiglböck, H.~Grosse, E.~H. Lieb, and
  W.~Thirring, Eds.\hskip 1em plus 0.5em minus 0.4em\relax Springer-Verlag,
  1993.

\bibitem{VonNeumann}
J.~V. Neumann, \emph{Mathematical Foundations of Quantum Mechanics (English
  Translation from )}, E.~Wigner and R.~Hofstader, Eds.\hskip 1em plus 0.5em
  minus 0.4em\relax Princeton University Press, 1955.

\bibitem{MMosonyi11}
M.~Mosonyi and F.~Hiai, ``On the quantum rényi relative entropies and related
  capacity formulas,'' \emph{IEEE Transactions on Information Theory}, vol.~57,
  no.~4, pp. 2474--2487, April 2011.

\bibitem{RBhatia}
R.~Bhatia, \emph{Matrix Analysis}, ser. Graduate Texts in Mathematics.\hskip
  1em plus 0.5em minus 0.4em\relax Springer, 1996.

\bibitem{ATeixeira12}
A.~Teixeira, A.~Matos, and L.~Antunes, ``Conditional rényi entropies,''
  \emph{IEEE Transactions on Information Theory}, vol.~58, no.~7, pp.
  4273--4277, July 2012.

\bibitem{RBhatia06}
R.~Bhatia, ``Infinite divisible matrices,'' \emph{The American Mathematical
  Monthly}, vol. 113, no.~3, pp. 221--235, March 2006.

\bibitem{RHorn69}
R.~A. Horn, ``The theory of infinitely divisible matrices and kernels,''
  \emph{Transactions of the American Mathematical Society}, vol. 136, pp.
  269--286, February 1969.

\bibitem{CBerg}
C.~Berg, J.~P.~R. Christensen, and P.~Ressel, \emph{Harmonic Analysis on
  Semigroups: Theory of Positive Definite and Related Functions}, ser. Graduate
  Texts in Mathematics.\hskip 1em plus 0.5em minus 0.4em\relax Springer-Verlag,
  1984.

\bibitem{ISchoenberg38}
I.~J. Schoenberg, ``Metric spaces and positive definite functions,''
  \emph{Transactions of the American Mathematical Society}, vol.~44, no.~3, pp.
  522--536, November 1938.

\bibitem{AGretton08}
K.~Fukumizu, A.~Gretton, X.~Sun, and B.~Schölkopf, ``Kernel measures of
  conditional dependence,'' in \emph{NIPS 2007}, J.~Platt, Ed., 2008.

\bibitem{LRosasco10}
L.~Rosasco, M.~Belkin, and E.~D. Vito, ``On learning with integral operators,''
  \emph{Journal of Machine Learning Research}, vol.~11, pp. 905--934, February
  2010.

\bibitem{TKato87}
T.~Kato, ``Variation of discrete spectra,'' \emph{Communications in
  Mathematical Physics}, vol. 111, pp. 501--504, 1987.

\bibitem{CCortes12}
C.~Cortes, M.~Mohri, and A.~Rostamizadeh, ``Algorithms for learning kernels
  based on centered alignment,'' \emph{Journal of Machine Learning Research},
  vol.~13, pp. 795--828, March 2012.

\bibitem{LSanchez13}
L.~G. Sanchez-Giraldo and J.~C. Principe, ``Information theoretic learning with
  infinitely divisible kernels,'' in \emph{International Conference on Learning
  Representations}, 2013.

\bibitem{AGretton10}
A.~Gretton and L.~Györfi, ``Consistent nonparametric test of independence,''
  \emph{Journal of Machine Learning Research}, vol.~11, pp. 1391--1423, April
  2010.

\end{thebibliography}
\end{document}